\documentclass[12pt]{article}
\usepackage{amsmath}
\usepackage{times}
\usepackage{graphicx}
\usepackage{color}
\usepackage{multirow}
\usepackage[authoryear]{natbib}
\usepackage{rotating}
\usepackage{bbm}
\usepackage{latexsym}
\graphicspath{ {figures/} }
\usepackage{subcaption}
\usepackage{amsfonts}

\newcommand{\captionfonts}{\normalsize}

\makeatletter  
\long\def\@makecaption#1#2{%
  \vskip\abovecaptionskip
  \sbox\@tempboxa{{\captionfonts #1: #2}}%
  \ifdim \wd\@tempboxa >\hsize
    {\captionfonts #1: #2\par}
  \else
    \hbox to\hsize{\hfil\box\@tempboxa\hfil}%
  \fi
  \vskip\belowcaptionskip}
\makeatother   

\begin{document}

\hspace{13.9cm}1

\ \vspace{20mm}\\

{\LARGE Progressive Interpretation Synthesis: Interpreting Task Solving by Quantifying Previously Used and Unused Information}

\ \\
{\bf \large Zhengqi He $^{\displaystyle a}$} \footnote{Email: zhengqi.he@riken.jp}\\
{\bf \large Taro Toyoizumi $^{\displaystyle a, \displaystyle b}$} \footnote{Email: taro.toyoizumi@riken.jp}\\
{$^{\displaystyle a}$Lab for Neural Computation and Adaptation, RIKEN Center for Brain Science, Saitama, Japan.}\\
{$^{\displaystyle b}$Department of Mathematical Informatics, Graduate School of Information Science and Technology, the University of Tokyo, Tokyo, Japan.}\\
%


\thispagestyle{empty}
\markboth{}{NC instructions}
\ \vspace{-0mm}\\
%

\newpage
\begin{center} {\bf Abstract} \end{center}

A deep neural network is a good task solver, but it is difficult to make sense of its operation. People have different ideas about how to form the interpretation about its operation. We look at this problem from a new perspective where the interpretation of task solving is synthesized by quantifying how much and what previously unused information is exploited in addition to the information used to solve previous tasks. First, after learning several tasks, the network acquires several information partitions related to each task. We propose that the network, then, learns the minimal information partition that supplements previously learned information partitions to more accurately represent the input. This extra partition is associated with un-conceptualized information that has not been used in previous tasks. We manage to identify what un-conceptualized information is used and quantify the amount. To interpret how the network solves a new task, we quantify as meta-information how much information from each partition is extracted. We implement this framework with the variational information bottleneck technique. We test the framework with the MNIST and the CLEVR dataset. The framework is shown to be able to compose information partitions and synthesize experience-dependent interpretation in the form of meta-information. This system progressively improves the resolution of interpretation upon new experience by converting a part of the un-conceptualized information partition to a task-related partition. It can also provide a visual interpretation by imaging what is the part of previously un-conceptualized information that is needed to solve a new task.


\section{Introduction}

Deep neural networks (DNNs) have made great achievements in fields such as image recognition \citep{krizhevsky2017imagenet}, speech recognition \citep{hinton2012deep}, natural language processing \citep{vaswani2017attention}, and game-playing beyond human-level performance \citep{silver2016mastering}. On the other hand, DNNs are famous black-box models. They fail under certain circumstances, such as adversarial attack \citep{goodfellow2014explaining}. This motivates an increasing trend of research into  understanding of how DNNs solve tasks, or model interpretation. Later research also suggests better model interpretation can be useful to, for example, explanation about model behavior, knowledge-mining, ethics, and trust. \citep{doshi2017towards, lipton2018mythos}

People have proposed different approaches to proceed with model interpretation. Concerning the interpretation style, there are the post-hoc style, which tries to separate the model training step and model interpretation step, and the concurrent style, which aims simultaneously for task performance as well as interpretation \citep{lipton2018mythos}. As for the applicability of interpretation methods, there is the model-specific type, targeting a certain class of models, and the model-agnostic type, where the interpretation method doesn't depend on the model \citep{arrieta2020explainable}. Considering the interpretation scope, there are the global interpretation and the local interpretation, where the global interpretation gives information about how the task is solved from a broader view while the local interpretation is more focused on certain examples or parts of the model \citep{doshi2017towards}. There are also diverse forms of interpretation, such as information feature \citep{chen2018learning}, relevance feature \citep{bach2015pixel}, a hot spot of attention \citep{hudson2018compositional}, or gradient information \citep{sundararajan2017axiomatic}. Another stream of research proposes that interpretable model are usually simple models like discrete-state models \citep{hou2018learning}, shallower decision trees \citep{freitas2014comprehensible, wu2017beyond}, graph models \citep{zhang2017interpreting}, or a small number of neurons \citep{lechner2020neural}. The readers can refer to \cite{arrieta2020explainable} for a more complete overview.

One particular dimension for model interpretation related to our paper is how much pre-established human knowledge is needed. Methods that require high human involvement, such as interpretation with human pre-defined concepts \citep{koh2020concept, chen2020concept} or with large human-annotated datasets \citep{kim2018visual}, implicitly assume the background knowledge of an average human to make sense of the interpretation, which is hard to be defined rigorously. Contrarily, existing human-agnostic methods transfer interpretation into some measurable form such as the depth of the decision tree \citep{freitas2014comprehensible, wu2017beyond}. However, how well this kind of measure is related to human-style interpretation is under debate. 

Within the human-agnostic dimension of interpretation, we extend the discussion along with two new perspectives. One perspective starts with the simple idea that ``interpretation should be experience-dependent". Motivated by this idea, we focus on the situation where the model learns a sequence of tasks by assuming that later tasks can be explained using earlier experiences. In other words, model interpretation in our framework is defined as meta-information describing how the information used to solve the new task is related to previous ones. The second perspective is motivated by the idea that ``interpretation should be able to handle the out-of-experience situation". In a situation where a new task cannot be fully solved by experience, the model interpretation method should be able to report new knowledge, mimicking a human explaining what is newly learned. We demonstrate this framework can cast insight onto how later tasks can be solved based on previous experience on MNIST and CLEVR datasets \citep{johnson2017clevr} and express ignorance when experience is not applicable.

Our work is related to the Concept Bottleneck Model (CBM) and Concept Whitening Model (CWM) \citep{koh2020concept,chen2020concept} in the sense that meaningful interpretation of the current task depends on previously learned knowledge. However, these methods do not capture reasonable interpretation when the human-defined concepts alone are insufficient to solve downstream tasks \citep{margeloiu2021concept}. In our framework, we add the un-conceptualized region to take care of information not yet associated with tasks. Moreover, a recent study also shows that contamination of concept-irrelevant information in the pre-defined feature space can hamper interpretation \citep{mahinpei2021promises}. We implement Information Bottleneck (IB) \citep{tishby2000information} as a remedy to this information leak problem. Our method also shares similarities with Variational Information Bottleneck for Interpretation (VIBI) method \citep{bang2019explaining} and the Multi-view Information Bottleneck method \citep{wang2019deep} in the sense that these methods use IB to obtain minimal latent representation from previously given representations. However, unlike the multi-view IB method for problem-solving, the goal of our framework is to synthesize interpretation. Furthermore, our framework does so using macroscopic task-level representations, which is different from microscopic input-level representations used in VIBI. 

\section{Intuitions}

This section discusses the intuition behind our framework for model interpretation.

\subsection{Interpretation as Meta-Information}

To quantify how a new task is solved using the experience of previous tasks, we evaluate meta-information. We define meta-information as a vector of mutual information, where each element of the vector describes how much the corresponding information partition is used for the new task.

\textbf{Interpretation in right level:} In this work, a machine learns a series of different tasks. The aim is to ascribe an interpretation of how the model solves the new task based on previous experience. If we did this using low-level features, such as the intensity and color of each pixel, the task description would become complicated. Instead, we aim to give an interpretation at a more abstract level, for example, ``This new task is solved by combining the knowledge about tasks 2 and 4." To achieve this goal, information about the input is partitioned at the task level. We, therefore, prepare information partitions that encode useful features for each task.

\textbf{Inducing independence:} So what conditions do these partitions have to satisfy? If these information partitions are redundant, we will have arbitrariness in assigning meta-information since a task can equally be solved using different partitions \citep{wibral2017partial}. Therefore, to have unambiguous meta-information, inducing independence among partitions is preferred. Useful methods are widely available in machine learning fields such as independent component analysis \citep{bell1995information, hyvarinen2000independent} and variational auto-encoders \citep{kingma2013auto}.

\textbf{Meaning assignment:} As described above, the meta-information we defined is a vector of Shannon information measured in bits (i.e., how much each information partition is used). While the number of bits itself doesn't have any meaning, each entry of the vector is linked to a corresponding task. Hence, the meta-information can be mapped to the relevance of previous tasks.

\subsection{Progressive Nature of Interpretation}

\textbf{Progressive interpretation:} One important but usually ignored property of interpretation is that we interpret based on experience \citep{national2002learning,bada2015constructivism}. Progressively learning multiple tasks is not a rare setting in machine learning \citep{andreas2016neural,rusu2016progressive,parisi2019continual}, which is usually named ``lifelong learning", ``sequential learning" or ``incremental learning". However, these studies usually focus on avoiding catastrophic forgetting and do not investigate how progressiveness contributes to interpretation. In one example \citep{kim2018visual}, the authors point out that interpretability emerges when lower-level modules are progressively made use of. We propose that interpretation should be synthesized in a progressive manner, where the model behavior is interpreted by how much the current task is related to previously experienced tasks.

\textbf{Knowing you don't know:} An experience-based progressive interpretation framework may inevitably encounter the situation when its previous experience does not help interpret the current task. To solve this problem, we introduce an ``un-conceptualized partition" storing information not yet included in the existing information partitions. We noticed that this un-conceptualized partition generates a ``knowing you don't know" type of interpretation, a meta-cognition ability that allows a person to reflect on their knowledge, including what they don't know \citep{glucksberg1981decisions}. Under this situation, the design of the framework should be able to interpret ``knowing you don't know" when faced with out-of-experience tasks.

We will formalize the intuitions in the language of information theory in the following sections.

\section{The Progressive Interpretation Framework}\label{sec3}

Assume we have a model with stochastic input $X$, which is statistically the same regardless of a task. Task $i$ is defined as predicting a series of stochastic labels $Z_i$. Its corresponding internal representation is $Y_i$. The progressive interpretation framework is formalized as iteratively as follows:
 
\begin{enumerate}

\item Assume after task $n$, a model has a minimal internal representation $Y=\{Y_1,Y_2, \\ \dots,Y_n,Y_{\rm else}\}$ that encodes the input $X$. $Y_i$ describes the internal representation learnt to solve task $i$. The optimization in the ideal case yields independence among the previous task-relevant partitions:
$$ I (Y_i; Y_j) = 0, (i\ne j \in[1,n]\cup{\rm else}). $$

Here, we define the notation $[1,n]$ to be $\{1,2,3,...,n\}$. 

\item Then, the model is faced with the new task $n+1$, and learns to predict $Z_{n+1}$. After learning $Z_{n+1}$, the model distills the necessary part $Y_{(i \cap n+1)}$ from each partition $Y_i (i=[1,n]\cup {\rm else})$ for solving task $n+1$. This is achieved by minimizing
$$ I(Y_{(i \cap n+1)};Y_i), (i \in[1,n]\cup{\rm else})$$
while maintaining the best task performance, i.e., by maintaining ideally all the task relevant information:
$$ I(\cup_{i=1}^{n, {\rm else}}{Y_{i}}; Z_{n+1}) = I(\cup_{i=1}^{n, {\rm else}}{Y_{(i \cap n+1)}}; Z_{n+1}).$$

\item The interpretation is defined as the meta-information of how much the individual partitions $\{Y_i\}$ for previous tasks $i \in [1,n]\cup{\rm else}$ are utilized to solve task $n+1$. Namely, the composition of the mutual information $I(Y_{(i \cap n+1)}; Y_i)$ over the different partitions $i=[1,n]\cup{\rm else}$ is the meta-information we use to interpret the global operation of the neural network. Then, local interpretation for each example is available from  $\{Y_{(i \cap n+1)}\}$.

\item After task $n+1$, the model updates the representation partition by splitting $Y_{\rm else}$ into the newly added representation $Y_{({\rm else} \cap n+1)}$ and its complement $Y_{\rm else}\setminus Y_{({\rm else} \cap n+1)}$. 
Then, the former is denoted as $Y_{n+1}$, and the latter is denoted as new $Y_{{\rm else}}$. 
The model would continue this for further iteration and interpretation of the tasks. 

\end{enumerate}

\begin{figure}[t]
\centering
\includegraphics[width=0.98\linewidth]{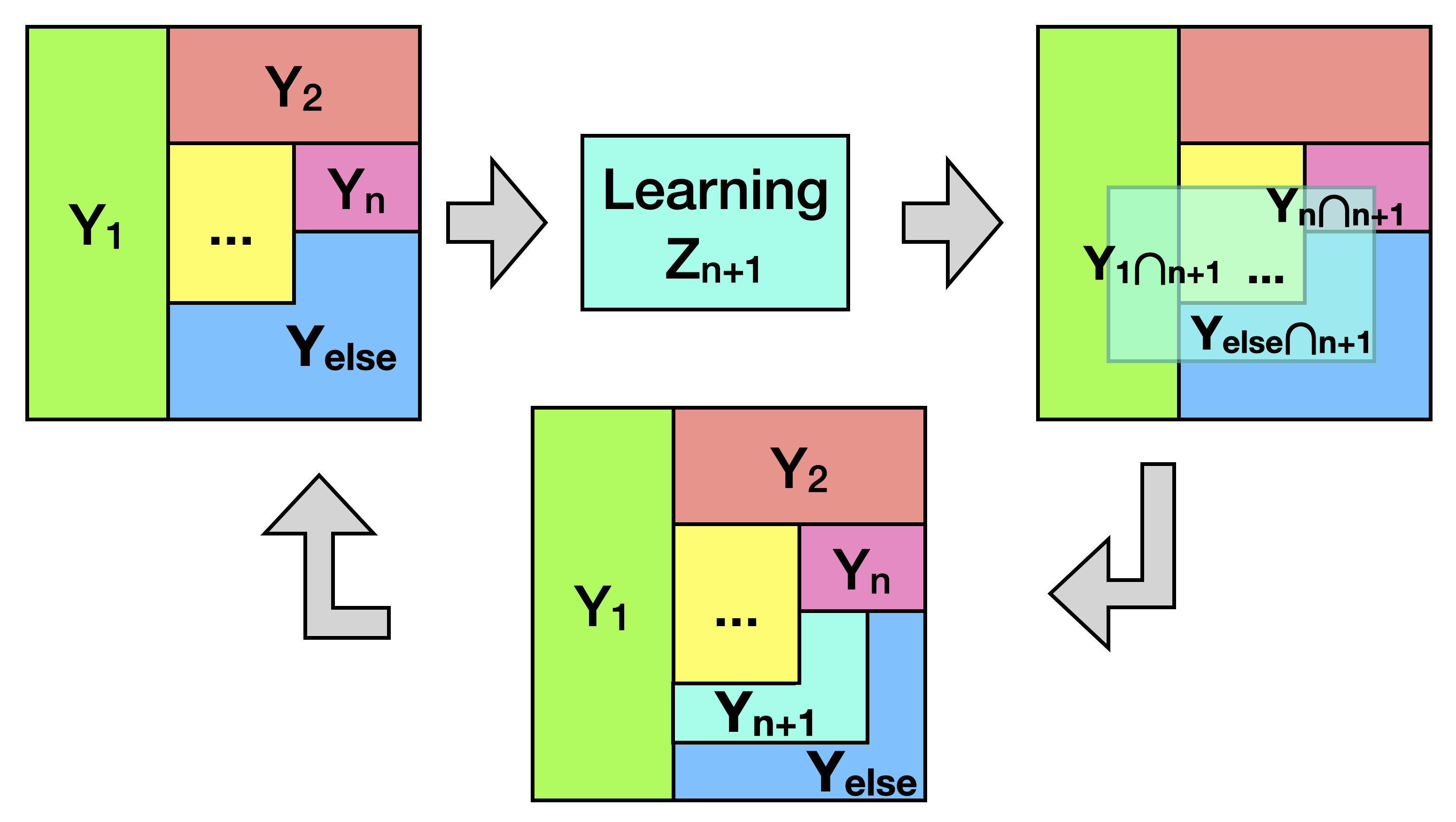}
\caption{A schematic plot showing the intuition of the progressive interpretation framework. Interpretation in our framework is based on the meta-information that specifies from which partitions the needed information comes to solve a new task $Z_{n+1}$. The map has the resolution in the level of task partitions $Y_i$, where partitions are made independent of each other. Independence among task partitions ensures the uniqueness of the needed information. Anything the model has not yet learned to utilize would stay in the un-conceptualized $Y_{else}$ region. The more tasks the model has encountered, the smaller the un-conceptualized region would be. Thus, a better interpretation can be gained for later tasks.}
\label{infograph}
\end{figure}

The intuition of the process is shown in Fig. \ref{infograph}. 

\section{Implementation} \label{sec4}

Our particular interest is in the system involving neural networks. Since our framework is information-theoretic, all types of neural networks are treated equally as segments of information processing pipelines. Which type of neural network to choose is decided by the specific problem.

\begin{figure}[t]
     \centering
     \begin{subfigure}[t]{1.0\linewidth}
        \centering
         \includegraphics[width=0.8\linewidth]{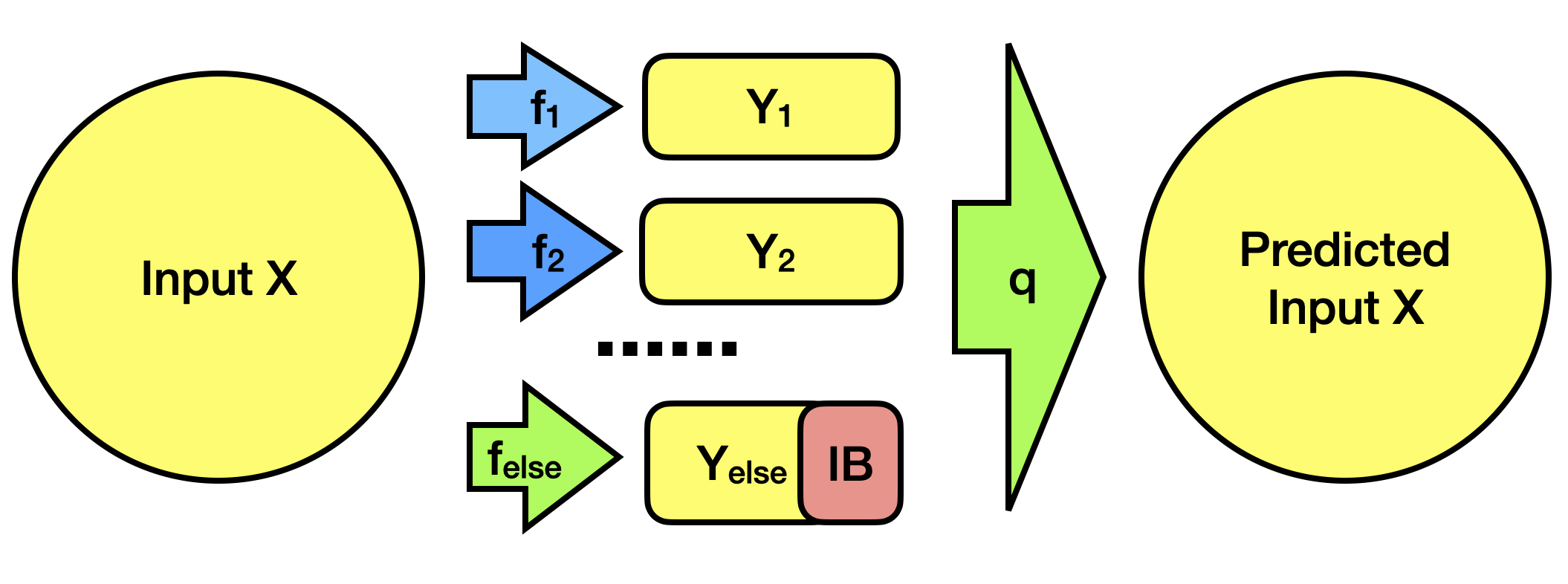}
         \caption{Information flow graph for information partition splitting of step 1.}
         \label{infoflowGa}
     \end{subfigure}
     \quad
     \begin{subfigure}[t]{1.0\linewidth}
        \centering
         \includegraphics[width=0.82\linewidth]{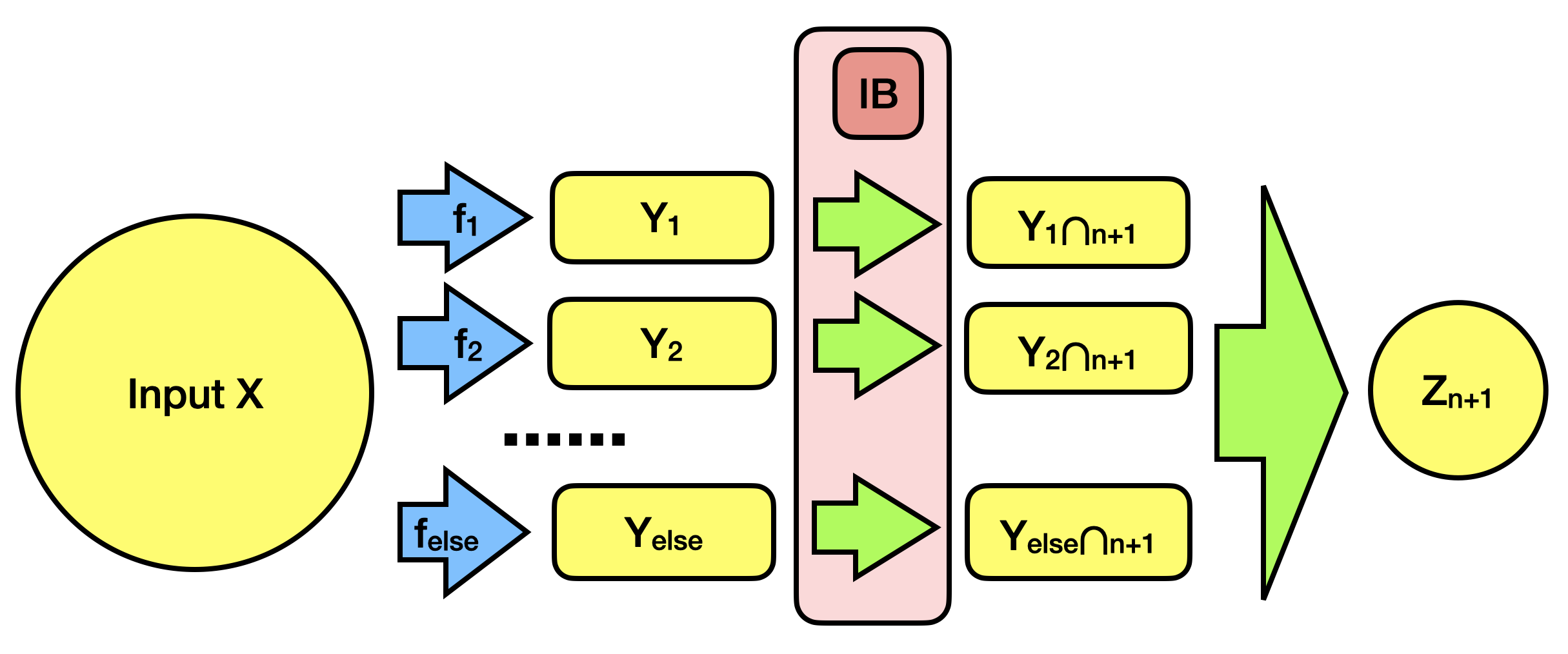}
         \caption{Information flow graph for interpretation of step 2.}
         \label{infoflowGb}
     \end{subfigure}
        \caption{Information flow graph of the progressive interpretation framework. Yellow areas are representations, and green and blue arrows represent neural networks. Green ones are put under training while blue ones are fixed. Red square with IB represents the information bottleneck.}
        \label{infoflowG}
\end{figure}

Neural network implementation of progressive interpretation can be implemented as loops over the four steps described in Section \ref{sec3}. In step 1, we assume a network already has information maps for task 1-to-$n$. After that, we extract the un-conceptualized partition that is unrelated to task 1-to-n by IB. In step 2, the model learns a new task $n+1$. Then, interpretation is gained by knowing how much information is needed from each sub-region as in step 3. In step 4, we repeat step 1 with a new map for task $n+1$ and prepare for the next loop. By adding new tasks and looping over the steps, a progressively more informative interpretation can be gained. The information flowing graph to implement in the following sections is shown in Fig. \ref{infoflowG}. 

\subsection{Information Bottleneck}

In our framework, IB plays an important role in manipulating information flow. To predict label $Z$ from statistical input $X$ with inner representation $Y$, IB would maximize:
\begin{equation} \label{infoB}
\max_\theta[{I(Y; Z)-\gamma I(Y;X)}], \; Y = f_{\theta}(X, \epsilon)
\end{equation} 
where $\gamma \in [0,1]$ is the scaling factor controlling the balance between the task performance (when $\gamma$ is small) and having non-redundant information representation (when $\gamma$ is large). $f$ is a neural network parameterized by the parameter $\theta$, and $\epsilon$ is a noise term that is important to suppress task-irrelevant information out of $X$.

We choose the variational information bottleneck (VIB) implementation \citep{alemi2016deep, chalk2016relevant, li2019specializing} with loss function
\begin{equation} \label{VinfoB}
L(p,q,r) = \mathbb{E}_{Y,Z} \left[-\log q\left(Z \mid Y \right)\right] + \gamma \mathbb{E}_X \{ \mathrm{KL}\left[p\left(Y \mid X\right), r(Y)\right] \}
\end{equation} 
to optimize encoding distribution $p(Y|X)$, decoding distribution $q(Z|Y)$, and the prior distribution $r(Y)$ for $p$. $\mathbb{E}_X$ describes taking the expectation over random variable $X$. Note that $\mathbb{E}_{Y,Z}=\mathbb{E}_X\mathbb{E}_{Y|X}\mathbb{E}_{Z|X}$. During the optimization, $\mathbb{E}_X\mathbb{E}_{Z|X}$ is computed by averaging over $N$ training samples of input $\{x_j| j=1,\dots,N\}$ and label $\{z_j| j=1,\dots,N\}$. $\mathbb{E}_{Y|X}$ is the average over the encoding distribution $p(Y|X)$, which is computed using the mapping $Y=f_{\theta}(X, \epsilon)$ of the encoding neural network. $Y$ can be a vector of either continuous or discrete variables \citep{li2019specializing} (see appendix Section \ref{sec:8-3} for details). For clarity, we further simplify the notation of loss function to be
\begin{equation} \label{VINsimp}
L = Q(Z|Y) + \gamma {\rm KL}(Y)
\end{equation} 
for future use, where the $Q$ term corresponds to the log-likelihood term trying to approximate $Z$ from internal representation $Y$. The ${\rm KL}$ term corresponds to the KL-divergence term trying to control the expressiveness of $Y$. 

\subsection{Task Training and Information Partition Splitting}
\label{sec:T1TIMP}

Suppose a new model with task input $X$ learns its first task to predict label $Z_1$. It is not difficult to train a neural network for this task by optimization:
$ \min_{\theta} D(f_{1,\theta}(X) || Z_1)$, 
where $D$ is a distance function, such as KL-divergence or mean-square error, which is decided by the problem. $f_{1,\theta}$ is an encoder network parameterized by $\theta$. After training, we will be able to obtain the representation of task 1 as $Y_1 = f_{1}(X, \epsilon)$, where $f_1$ indicates a neural network$f_{1,\theta}$ after optimizing $\theta$. 

Then, our next problem is how to obtain task 1 unrelated representation $Y_{\rm else}$, which ideally satisfies $I(Y_1; Y_{\rm else})= 0$, to complement the intermediate representation about the input. Here, we propose that $Y_{\rm else}$ can be obtained via the implementation of IB on an auto-encoding task:
\begin{equation} \label{infoB}
\begin{split}
& \max_\theta[{I(Y_1,Y_{\rm else}; X)-\gamma I(Y_{\rm else};X)}], \\
& Y_{\rm else}=f_{{\rm else}, \theta}(X, \epsilon),
\end{split}
\end{equation} 
where $\gamma$ is again the scaling factor controlling the trade-off between including and excluding different information. Note that the learned $f_1$ function is fixed while $f_{{\rm else},\theta}$ is trained. The intuition behind Eq. \ref{infoB} is described as follows. $I(Y_1; Y_{\rm else}) > 0$ implies redundant information about $Y_{1}$ contained in $Y_{else}$. This redundant information wouldn't improve $I(Y_1,Y_{\rm else}; X)$. However, removing this redundant information can decrease $I(Y_{\rm else};X)$, thus contributing to our optimization goal. Note that we assume $\gamma$ is less than one.

With the simplified notation of the VIB introduced above, the loss function
\begin{equation} \label{VIBSplit}
L = Q(X|Y_{1},Y_{\rm else}) + \gamma {\rm KL}(Y_{\rm else})
\end{equation} 
is minimized. The loss function seeks to auto-encode $X$ given previously learned $Y_1$ (which is fixed) together with $Y_{\rm else}$, while controlling expressiveness of $Y_{\rm else}$.

\subsection{New Task Interpretation}
\label{sec:T2MC}

Now assume the model has internal representation $ Y=\{Y_1, Y_2, ..., Y_n, Y_{\rm else}\} $ after learning tasks 1 to $n$. When the new task $n+1$ is introduced, the model learns to predict $Z_{n+1}$. Task $n+1$ relevant information can be extracted from $Y$ by the IB as follows:
\begin{equation} \label{interpEq}
\begin{split}
& \max_{\theta}\left [I(\cup_{i=1}^{n, {\rm else}}{Y_{(i \cap n+1)}}; Z_{n+1})-\gamma \sum_{i=1}^{n,{\rm else}} I(Y_{(i \cap n+1)};Y_i)\right ], \\
& Y_{(i \cap n+1)} = f_{(i \cap n+1),\theta}(Y_i, \epsilon)
\end{split}
\end{equation} 
where $Y_{(i \cap n+1)}, (i \in[1,n]\cup{\rm else})$ is the information needed from $Y_i$ to solve task $n+1$. Again, $\epsilon$ is the noise term required to eliminate information irrelevant to task $n+1$. Since $Y_{(i \cap n+1)} = f_{(i \cap n+1),\theta}(Y_i, \epsilon)$ depends on $Y_i$, together with IB, $Y_{(i \cap n+1)}$ is then a minimum sub-partition of $Y_i$ required for task $n+1$. 
We again implement the variational IB loss function with simplified notation:
\begin{equation} \label{VinfoB2}
L = Q(Z_{n+1}|\cup_{i=1}^{n, {\rm else}}{Y_{(i \cap n+1)}} ) +\frac{\gamma}{n+1} \sum_{i=1}^{n,{\rm else}} {\rm KL}(Y_{(i \cap n+1)})
\end{equation} 
The loss function seeks to maximize the prediction of $Z_{n+1}$ while controlling the needed information from $Y_i$. Index $i$ specifies a representation partition. 

After getting  $\{Y_{(i \cap n+1)}\}$, we can get interpretation as the meta-information $I(Y_{(i \cap n+1)}; Y_i)$ needed from each partition $Y_i$ as defined in Sec. \ref{sec3}. We can also look into the representations of $Y_{(i \cap n+1)}$ to gain insight about how task $n+1$ is solved for each example.

$Y_{({\rm else} \cap n+1)}$ is the information needed from the un-conceptualized partition $Y_{\rm else}$ to solve task $n+1$. We can rewrite this to be $Y_{n+1}$ and define the new un-conceptualized partition as $Y_{{\rm else}} \gets Y_{\rm else}~\setminus~Y_{({\rm else};n+1)}$. We can then go back to Step 1 and continue the iteration for task $n+2$.

\section{Experiments} \label{sec5}

\subsection{MNIST Dataset}

We first illustrate our progressive interpretation framework on the MNIST dataset (60K/10K train/test splits). We set task 1 as digit recognition. For task 2, we propose three kinds of tasks: telling if a number is even or odd (parity task), predicting the sum of pixel intensities (ink task), or a task that involves both digit information and pixel intensity information with a certain resolution (see below). First, we train a network $f_1$ to perform digit recognition, and then we train an auto-encoder with IB to train a network $f_{\rm else}$ to obtain a digit-independent partition. Then, we extend the network to train on a second task and obtain interpretation from the information flow. We choose continuous latent representation for this section. See appendix Sec.\ref{sec:8-1} \ref{sec:8-2} for implementation details.

\subsubsection{IB Removes Task-relevant Information from the Un-conceptualized Region}

\begin{figure}[t]
     \centering
     \begin{subfigure}[t]{0.49\linewidth}
         \centering
         \includegraphics[width=1.0\linewidth]{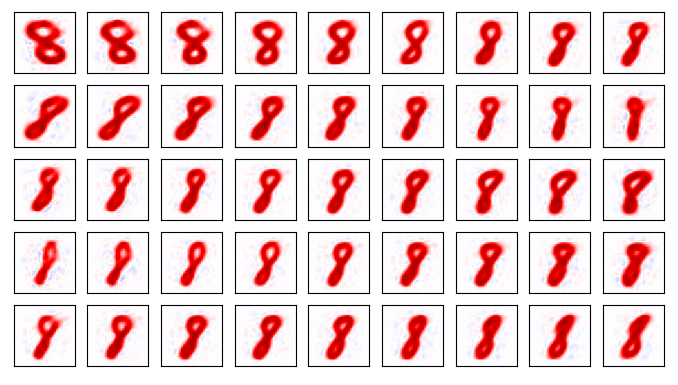}
         \caption{$Y_{\rm else}$ Before ink task.}
         \label{number_remove}
     \end{subfigure}
     \begin{subfigure}[t]{0.49\linewidth}
         \centering
         \includegraphics[width=1.0\linewidth]{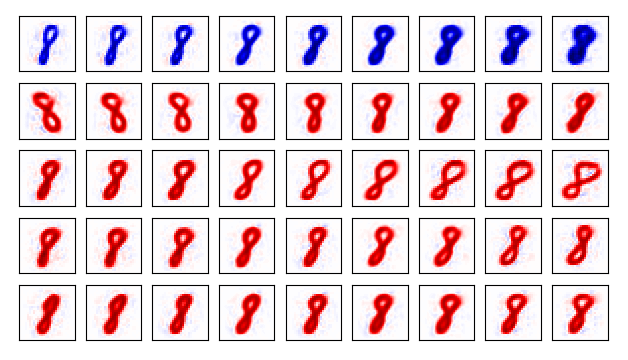}
         \caption{$Y_{\rm ink}$ (first row) and $Y_{\rm else}$ after ink task.}
         \label{task2_auto}
     \end{subfigure}
     \quad
     \begin{subfigure}[t]{1.0\linewidth}
         \centering
         \includegraphics[width=1.0\linewidth]{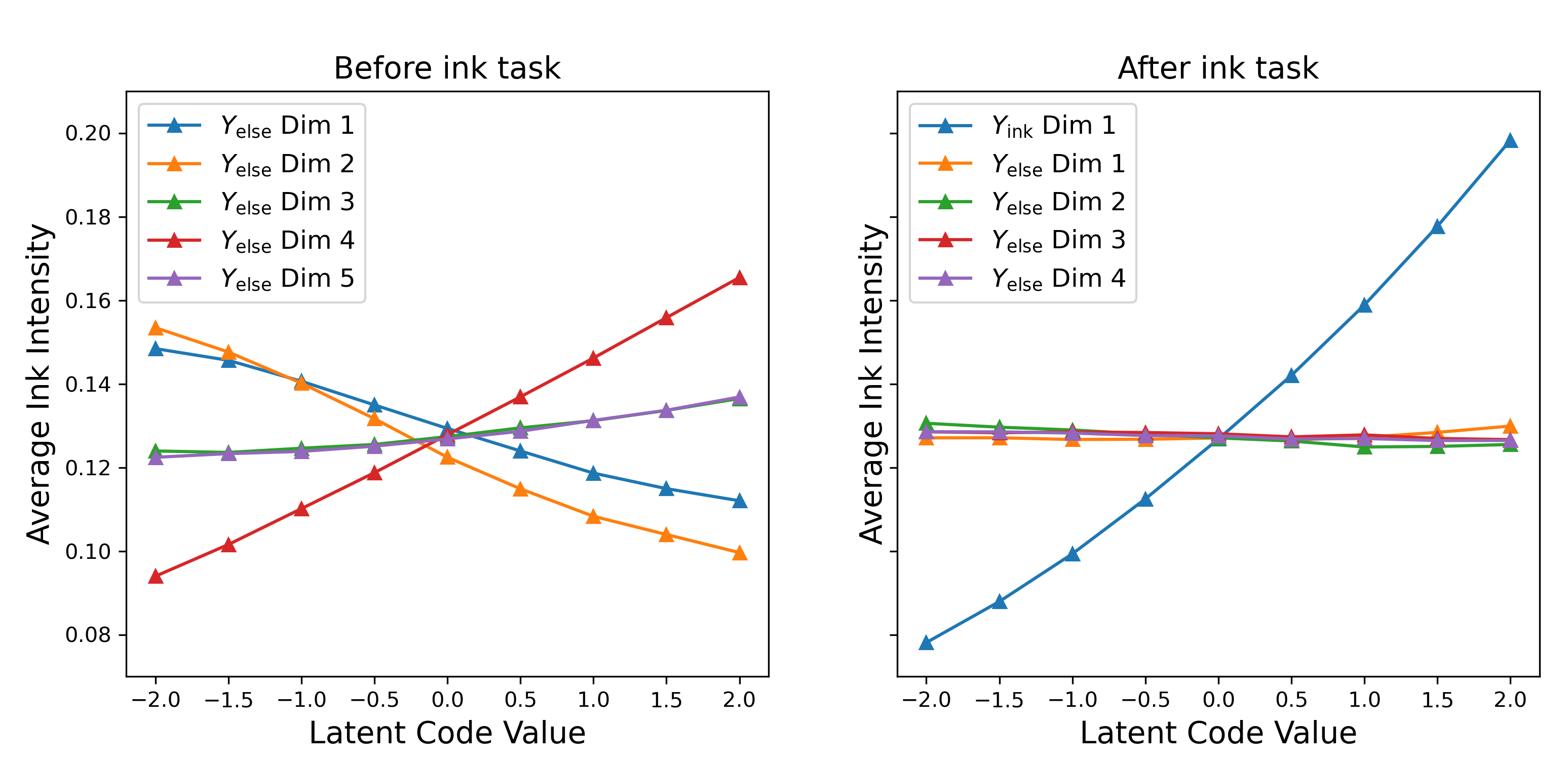}
         \caption{Average ink intensity v.s. latent code value of the five dimensions before and after the ink task.}
         \label{task2_scan}
     \end{subfigure}
        \caption{Latent code scanning of un-conceptualized representation after auto-encoding, before (a) and after (b) ink task (except the first row). The reconstructed images are plotted as the activity (columns) of one of the coding units (rows) is varied with others fixed. Fig. (c) shows how the average ink intensity varies when we scan the latent code of the same 5 units as in (a) and (b).}
        \label{lcodescan}
\end{figure}

Un-conceptualized representation can be obtained after the auto-encoding step. We can check what has been learned by scanning this latent code. Fig. \ref{number_remove} shows the scanning result of the top five latent representation units, ordered by descending mutual information with $X$. Note that changing these features doesn't change the digit. Moreover, mutual information between $Y_{\rm digit}$ and $Y_{\rm else}$ is estimated by training a neural network that predicts $Y_{\rm digit}$ from $Y_{\rm else}$. The estimated information is smaller than 0.1 Nat when $\gamma$ is larger than 5e-4, which indicates that digit information is removed from the un-conceptualized region by IB.

\subsubsection{The Framework Explains How a New Task is Solved}

\begin{figure}[t]
\centering
\includegraphics[width=0.98\linewidth]{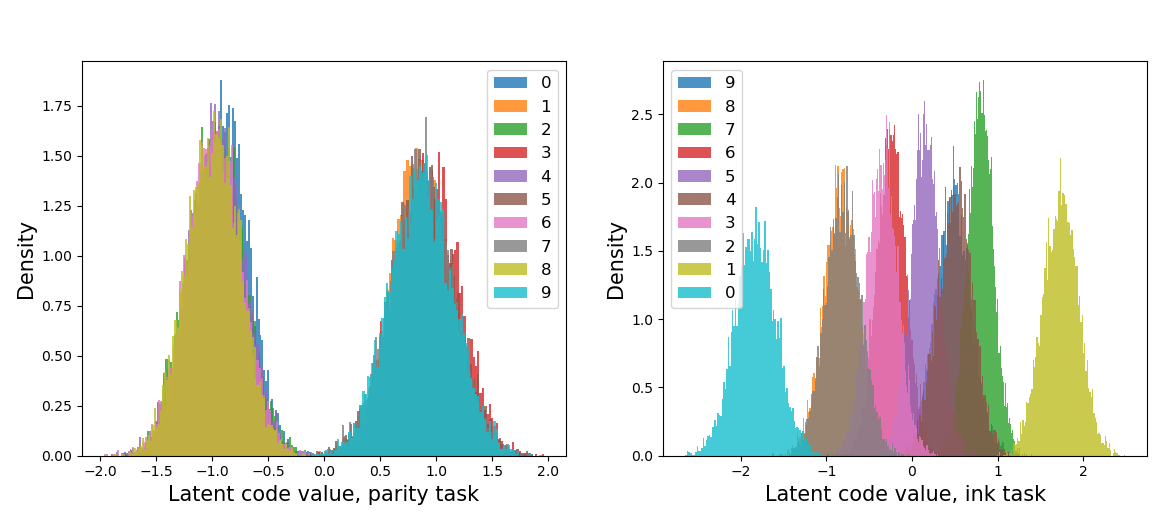}
\caption{VIB latent code distribution of different digits for the parity task $Y_{\rm digit \cap \rm parity}$ (left) and ink task $Y_{\rm digit \cap \rm ink}$ (right). X axis shows the code value, Y axis shows the code density, and different colors represent different digits ranging from 0 to 9. It can be seen that, for the parity task, the latent code formed two clusters, one for even numbers and one for odd numbers. And for the ink task, digits are aligned in the order of the average ink amount.}
\label{lcodeintp}
\end{figure}

After the auto-encoding step, we proceed to solve either the parity task or ink task to study the interpretation that the framework provides. For the parity task, mutual information from $Y_{digit}$ and from $Y_{else}$ are 0.702 Nat and 0.002 Nat respectively, and for the ink task, 1.498 Nat and 2.045 Nat. The result shows that the parity task doesn't need information from $Y_{else}$, while the ink task does. Clues of how the tasks are solved can also be found by looking into the representation obtained after IB. For the parity task, different digits are clustered into two groups according to their parity. For the ink task, digits are aligned in an order corresponding to their actual average ink amount ($0>8>2>3>6>5>9>4>7>1$) as Fig. \ref{lcodeintp} shows.

\subsubsection{Experience-dependence of the ELSE Partition}

After learning the digit task and, then, the ink task, we can update the auto-encoder $f_{\rm else}$ to exclude the ink-task-related information. On the one hand, $Y_{\rm ink}$ (first row of Fig. \ref{task2_auto}) represents the average pixel intensity. On the other hand, this information is suppressed in $Y_{\rm else}$ (rows 2-5). The suppression can be measured by feature correlation between $Y_{\rm ink}$ and $Y_{\rm else}$. Before ink task, the correlations are (0.295, 0.414, 0.080, 0.492, 0.100) for the 5 units visualized, but after the ink task, the correlation becomes (0.030, 0.194, 0.019, 0.028, 0.001). We also present the result of the average ink intensity v.s. latent code of the 5 units. It can clearly be seen that, before the ink task, the knowledge of average intensity is distributed across all 5 units. However, after the ink task, the knowledge of average intensity is extracted as $Y_{\rm ink}$ and removed from $Y_{\rm else}$ (Fig. \ref{task2_scan}). The result indicates that the un-conceptualized region is experience-dependent and information about the already learned task is excluded. Unlike other frameworks such as variational auto-encoder \citep{kingma2013auto} and infoGAN \citep{chen2016infogan} which usually have no explicit control over partitioning latent representation, our framework allows latent representation re-organization through progressive tasks.

\subsubsection{Quantitative benchmark of interpretation}
\label{sec:MNISTBench}

\begin{figure}[t]
\centering
\includegraphics[width=0.98\linewidth]{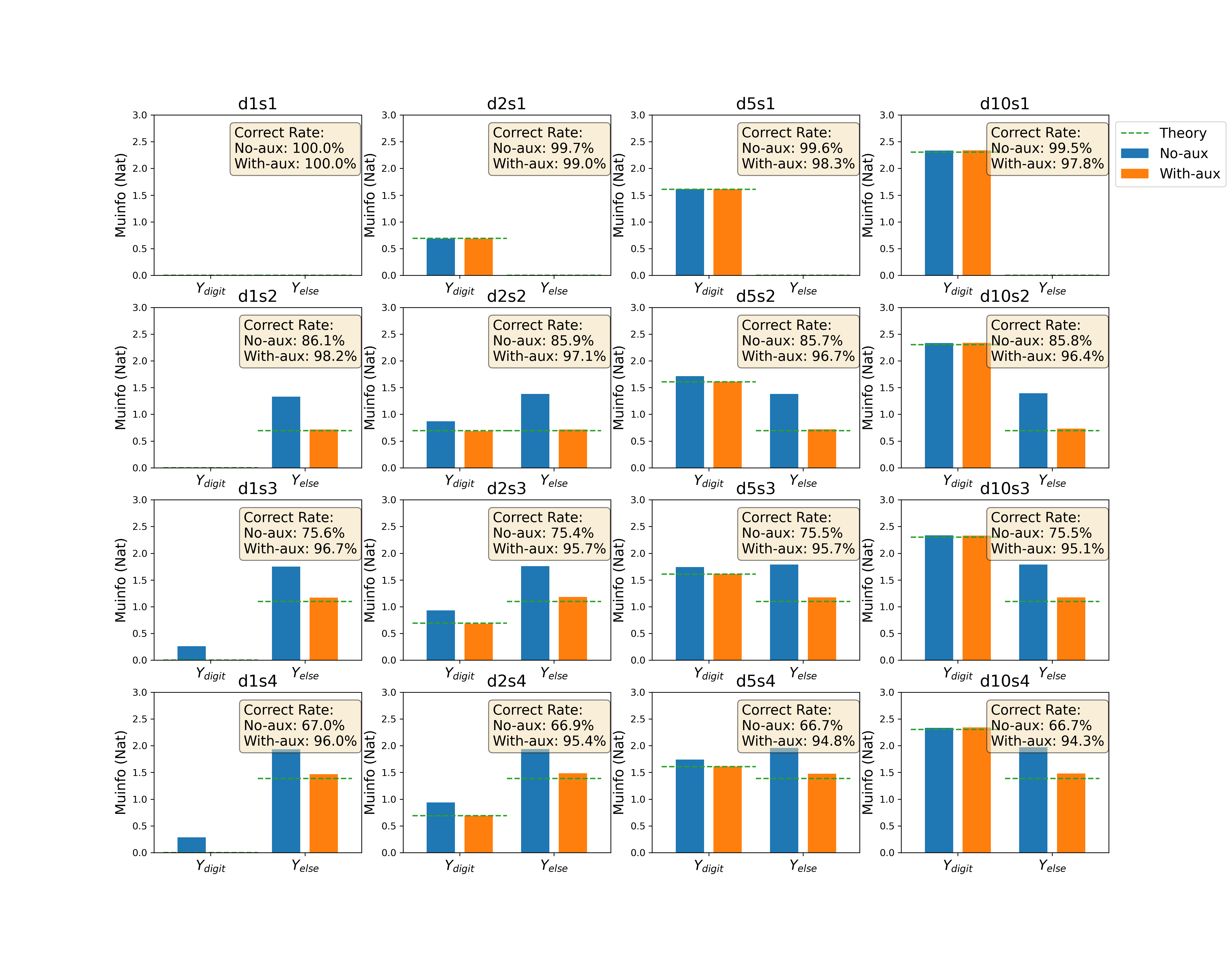}
\caption{Mutual information from $Y_{\rm digit}$ and $Y_{\rm else}$ used to solve the benchmark tasks. Blue/orange bars are mutual information without/with auxiliary auto-encoding, and the theory indicating the required amount of information is plotted with the green dotted line. Inside text boxes are task correct rates without/with auxiliary auto-encoding. The title of each subplot represents different task types combining 4 digit resolutions: d1, d2, d5, d10, and 4 digit-independent ink resolution : s1, s2, s3, s4, forming a 4-by-4 matrix.}
\label{taskmat_interp}
\end{figure}

After that, we ask if our proposed interpretation is quantitatively useful. Because we are not aware of task-level human-agnostic interpretation algorithms directly comparable to ours, we study how the interpretation changes as we systematically modify the required type of information for task 2. Task 2 is designed to require both digit information and digit-independent ink information involving different resolutions. For digit information, we have 4 resolutions: d1, d2, d5, d10. For example, d5 means 10 digits are separated into 5 equally sized groups and the task is to tell which group the image belongs to. As a result, (0, 0.693, 1.609, 2.303) Nat of information about digit is theoretically needed, respectively. For digit-independent ink information, we also have 4 resolutions (according to the percentile-based grouping for each digit by the amounts of ink used): s1, s2, s3, s4, which theoretically require (0, 0.693, 1.099, 1.386) Nat of information. By combining them, we get 16 different possibilities for task 2, and the interpretation measured as mutual information and the corresponding theoretical values are shown in Fig. \ref{taskmat_interp}. The figure shows that information needed from $Y_{\rm digit}$, $I(Y_{\rm digit\cap 2}; Y_{\rm digit})$, can be precisely predicted. The required non-digit information $I(Y_{\rm else \cap 2}; Y_{\rm else})$ from $Y_{\rm else}$ via auto-encoding correlates with the required amount to solve the task. However, due to the imperfection of the variational IB algorithm to purely extract relevant information, more than the theoretically required amount of information from $Y_{\rm else}$ is used for good performance. This problem can be practically remedied by allowing $Y_{\rm else}$ to be re-trained by adding an auxiliary auto-encoding task when learning task 2. Since input data is available during task 2, adding an auxiliary auto-encoding task during task 2 training increases task 2 performance without needing extra data. See appendix Section \ref{sec:8-9} for further discussion.

\subsection{CLEVR Dataset}

In this section, we demonstrate the progressive interpretation framework on the CLEVR dataset \citep{johnson2017clevr}. The CLEVR dataset is a large collection of 3D rendered scenes (70K/15K train/test splits) with multiple objects with compositionally different properties. The CLEVR dataset is originally designed for a visual question-answering task but we train the model without using natural language. For example, we train the model to classify the color of an object or conduct a multiple-choice (MC) task using only the pictures. For the multiple-choice task, the model is trained on a large set of four pictures and learns to choose one of the four pictures that includes a target object (100K/20K train/test splits).

In this section, we divide the tasks into two groups. Task group 1: the model that is pre-trained to tell objects apart learns to recognize part of the important properties among shape, size, color, material, and position. Task group 2: the model is asked to perform a multiple-choice task selecting a picture according to a specific context, for example, ``choose the picture with red cubes," which needs information learned or not yet learned in task 1. For task group 1, we first use convolutional neural networks (CNNs) to report the image properties by supervise learning and then obtain the un-conceptualized region via auto-encoding. After that, task group 2 is performed with interpretation synthesized. We choose discrete latent representation for this section. See appendix Section \ref{sec:8-1} \ref{sec:8-2} for Implementation details.

\subsubsection{Interpretation by Information Flow}

The result of interpretation by information flow is shown in Table \ref{info-table}. The mutual information $I(Y_{(i\cap{\rm MC})};Y_i)$ for $i\in\{{\rm posi}, {\rm color}, {\rm material}, \rm else\}$ is measured in Nat per object, where MC represents the multiple-choice task. Different rows represent different question types. We sample 5 random initializations of the networks for each task and present both the average and standard deviation. The theoretical amount of information required for feature $i$ is shown in parentheses. We can interpret how the model is solving the task by calculating mutual information coming from each information partition. For example, the task to ``choose the picture with a green metal" needs 0.345 Nat of information from the color domain and 0.686 Nat from the material domain. Information coming from other domains is judged as irrelevant to this task, which is as expected. If the task is ``choose the picture with a small yellow object," the model then needs 0.343 Nat from the color domain, plus 0.70 Nat of information from the un-conceptualized region since the model has not yet explicitly learned about using object size. If the task is ``choose the picture with a large sphere," the model finds out all previously learned properties are useless and has to pick 0.31 Nat of information from the un-conceptualized region. This is because neither size nor shape information has been used in previous tasks.

\begin{table*}[h]
\caption{Table for Task2 interpretation, information unit (Nat/object), inside parentheses is the theoretical value.}
\label{info-table}
\begin{center}
\resizebox{\textwidth}{!}{
\begin{tabular}{|l|ccccr|}\hline
Question Type&	Position &	Color &	Material &	Unknown &	Correct rate \\
\hline
Green Metal&	$<0.001$ (0) &	$0.345 \pm 0.001$ (0.377) &	$0.686 \pm 0.008$ (0.693)&	$<0.001$ (0) &$99.3 \pm 0.1\%$ \\
Left Rubber&	$0.56 \pm 0.02 (0.52)$&	$<0.001$ (0)&	$0.688 \pm 0.001$ (0.693) &	$0.01 \pm 0.02$ (0)&	$97.0 \pm 0.7\%$ \\
Small Yellow&	$<0.001$ (0)&	$0.343 \pm 0.002$ (0.377)&	$<0.001$ (0)&	$0.70 \pm 0.01$ (0.693)&	$99.2 \pm 0.1\%$ \\
Red Cube&	$<0.001$ (0)&	$0.381 \pm 0.002$ (0.377) &	$<0.001$ (0)&	$0.89 \pm 0.06$ (0.637)&	$95.8 \pm 0.4\%$ \\
Right Cylinder&	$0.59 \pm 0.03$ (0.51)&	$<0.001$ (0)&	$<0.001$ (0)&	$0.88 \pm 0.06$ (0.637)&	$94.8 \pm 0.7\%$ \\
Large Sphere&	$<0.001$ (0)&	$<0.001$ (0)&	$<0.001$ (0)& 	$0.31 \pm 0.01$ (0.451)&	$99.4 \pm 0.1\%$ \\
\hline
\end{tabular}}
\end{center}
\end{table*}

\subsubsection{Single Example Interpretation and Un-conceptualized Representation}

\begin{figure}[t]
\centering
\includegraphics[width=0.98\linewidth]{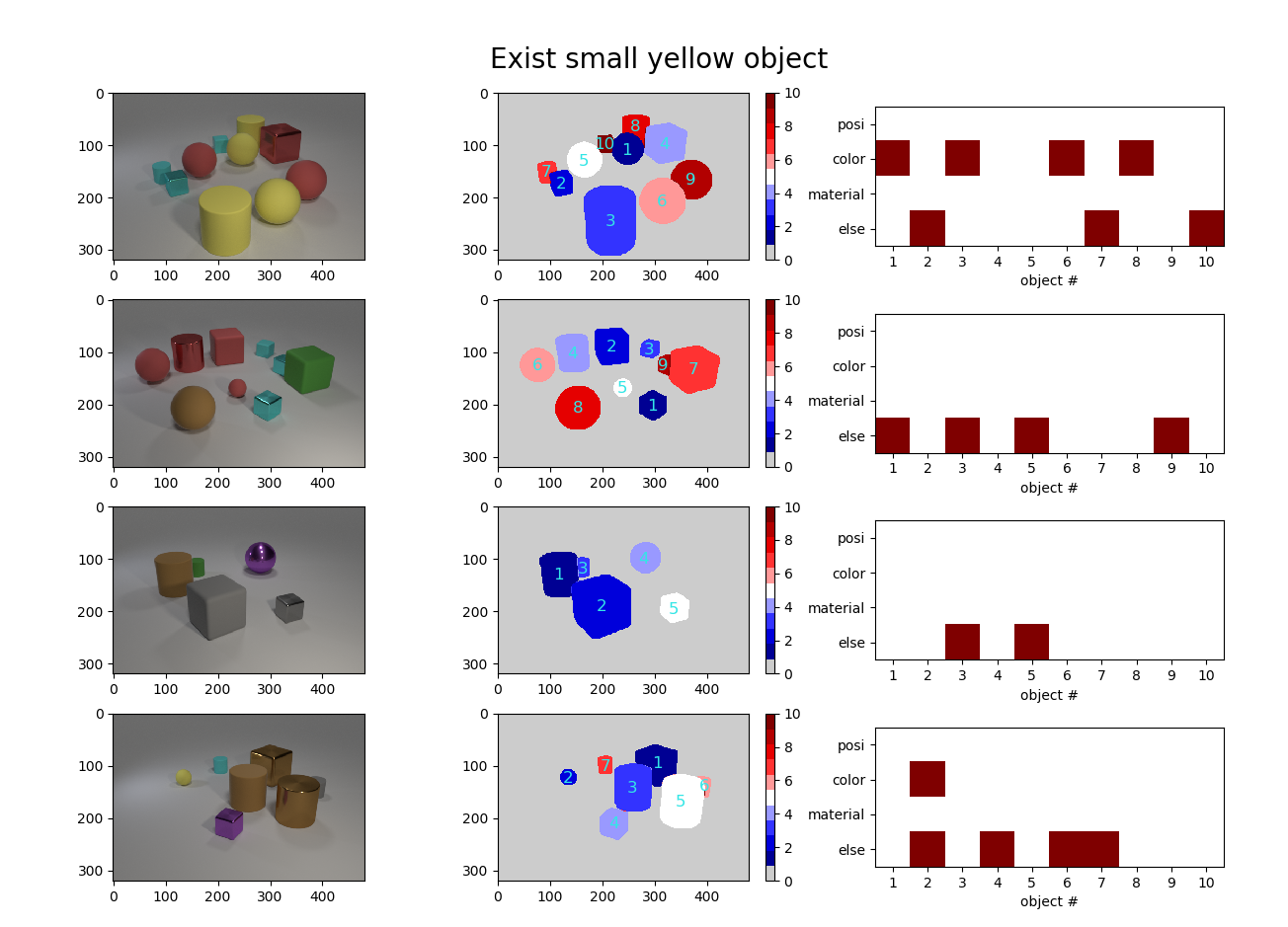}
\caption{Single-example interpretation of the task ``choose the small yellow object." The left column shows input pictures, and the middle column shows masks colored according to object IDs. We overlaid the masks with the object IDs for visual aid. The right column shows the binary activity summarizing the information at layer $Y_{(i \cap MC)}$. The x-axis corresponds to object ID, the y-axis represents four kinds of representations, namely position $Y_{({\rm posi} \cap {\rm MC})}$, color $Y_{({\rm color} \cap {\rm MC})}$, material $Y_{({\rm material} \cap {\rm MC})}$, and else $Y_{({\rm else} \cap {\rm MC})}$, where the dimension with highest mutual information is plotted. The red square represents the lower frequency binary representation, and the white represents the counterpart.}
\label{single}
\end{figure}

After getting the model, it is also possible to synthesize interpretation for a single example by looking into the discrete representation $Y_{(i\cap{\rm MC})}$ for $i\in\{{\rm posi}, {\rm color}, {\rm material}, \rm else\}$.  A typical example is shown in Fig. \ref{single}. This example corresponds to a ``small yellow object." We can see the model discriminates if the object has the color ``yellow" while neglecting position and material information. To solve the problem,  the model also needs information from the un-conceptualized partition which is representing the size ``small." The behavior of the model is consistent with the expectation of the question regarding the ``small yellow object."

We examine the correctness of the un-conceptualized representation by comparing it with the true label. For example, if the task is ``choose the small yellow object," the un-conceptualized region should represent the size ``small." We can cross-check by calculating their mutual information, which is 0.662 Nat per object. For the case ``choosing red cube", mutual information with the label ``cube" is 0.432 Nat per object. For the case ``choosing cylinder on the right side", mutual information with the label ``cylinder" is 0.408 Nat per object. All these numbers exceed the chance level (the 99, 95, and 90-percentile by chance are 0.637, 0.495, and 0.368 Nat respectively for balanced binary random variables like size, and 0.583, 0.449, 0.332 Nat for cases with three alternatives like shape).

\subsubsection{Visualizing the Un-conceptualized Representation}

After getting the un-conceptualized representation useful for the new task, we can continue the framework by splitting the un-conceptualized representation into the learned useful part and its complement. Separating this new useful representation is non-trivial because labels of the MC task jointly depend on multiple image properties. While previous methods \citep{koh2020concept,chen2020concept} need feature-specific labels to learn a new property, the proposed framework automatically segregates a new useful representation from previously learned representations. Furthermore, the proposed system can visualize what a new representation has just been learned.

\begin{figure}[t]
\centering
\includegraphics[width=0.98\linewidth]{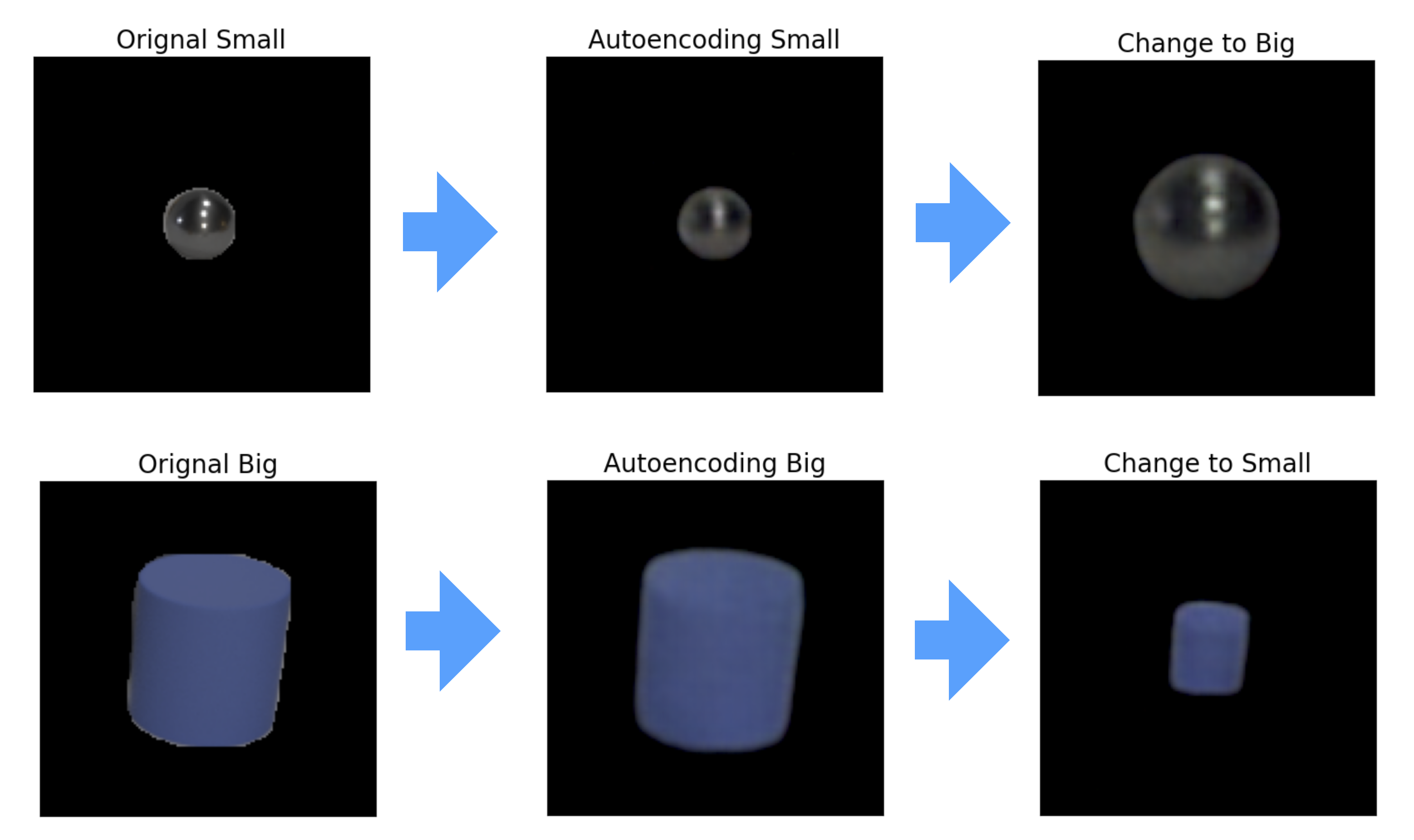}
\caption{Visualizing the newly learned $Y_{\rm MC}$ about size after learning the task ”choose the picture with a small yellow object.” As can be seen from the result, changing $Y_{\rm MC}$ of a small object renders a big counterpart of the same object, and changing $Y_{\rm MC}$ of a big object would render a small counterpart of the same object.}
\label{unconcept_gen}
\end{figure}

Here, we demonstrate the result after learning the task ”choose the picture with a small yellow object.” We mentioned above that after learning this new task, the model is expected to learn new concept about size as the new representation $Y_{\rm MC} = Y_{(\rm else \cap \rm MC)}$. Note, again, that we never provided the model labels specifically about size. Then we can continue the framework by performing another round of auto-encoding, which splits $Y_{\rm else}$ into $Y_{\rm MC}$ and $Y_{\rm else} \setminus Y_{\rm MC}$. After that, the model explains what property is newly learned by generating the image of an object and changing its size as the newly latent representation $Y_{\rm MC}$ is altered (Fig. \ref{unconcept_gen}). This visualization also helps humans interpret the operation of the model.

Information about other studies on the CLEVR dataset can be found in Section \ref{sec:8-4} to section \ref{sec::highorder}. We also perform more discussions about our method in appendix Section \ref{sec:8-9}, and discuss about limitations of our method in appendix \ref{sec:8-10}.

\section{Conclusion}

This paper proposes a progressive framework based-on information theory to synthesize interpretation. We show that interpretation involves independence, is progressive, and can be given at a macroscopic level using meta-information. Changing the receiver of the interpretation from a human to a target model helps define interpretation clearly. Our interpretation framework divides the input representations into independent partitions by tasks and synthesizes interpretation for the next task. This framework can also visualize what conceptualized and un-conceptualized partitions code by generating images. The framework is implemented with a VIB technique and is tested on the MNIST and the CLEVR dataset. The framework can solve the task and synthesize non-trivial interpretation in the form of meta-information. The framework is also shown to be able to progressively form meaningful new representation partitions. Our information-theoretic framework capable of forming quantifiable interpretations is expected to inspire future understanding-driven deep learning. 

\section{Acknowledgement}

We would like to thank Ho Ka Chan, Yuri Kinoshita and Qian-Yuan Tang for useful discussions about the work. This study was supported by Brain/MINDS from Japan Agency for Medical Research and Development (AMED) under Grant Number JP15dm0207001, Japan Society for the Promotion of Science (JSPS) KAKENHI Grant Number JP18H05432, and RIKEN Center for Brain Science.

\section{Appendix}

\subsection{Experiment Details}
\label{sec:8-1}

\subsubsection{MNIST Case}

\begin{figure}[t]
     \centering
     \begin{subfigure}[t]{0.98\linewidth}
         \centering
         \includegraphics[width=\textwidth]{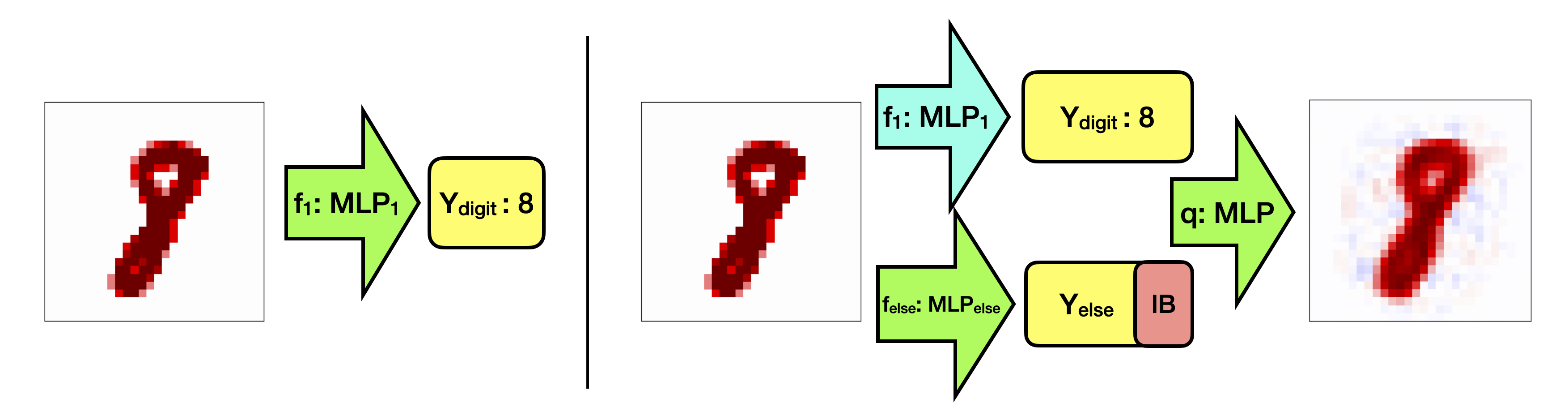}
         \caption{Left: digit supervised learning by $\mathrm{MLP}_{1}$. Right: information map splitting via auto-encoding.}
         \label{digitMLP}
     \end{subfigure}
     \begin{subfigure}[t]{0.98\linewidth}
         \centering
         \includegraphics[width=0.75\linewidth]{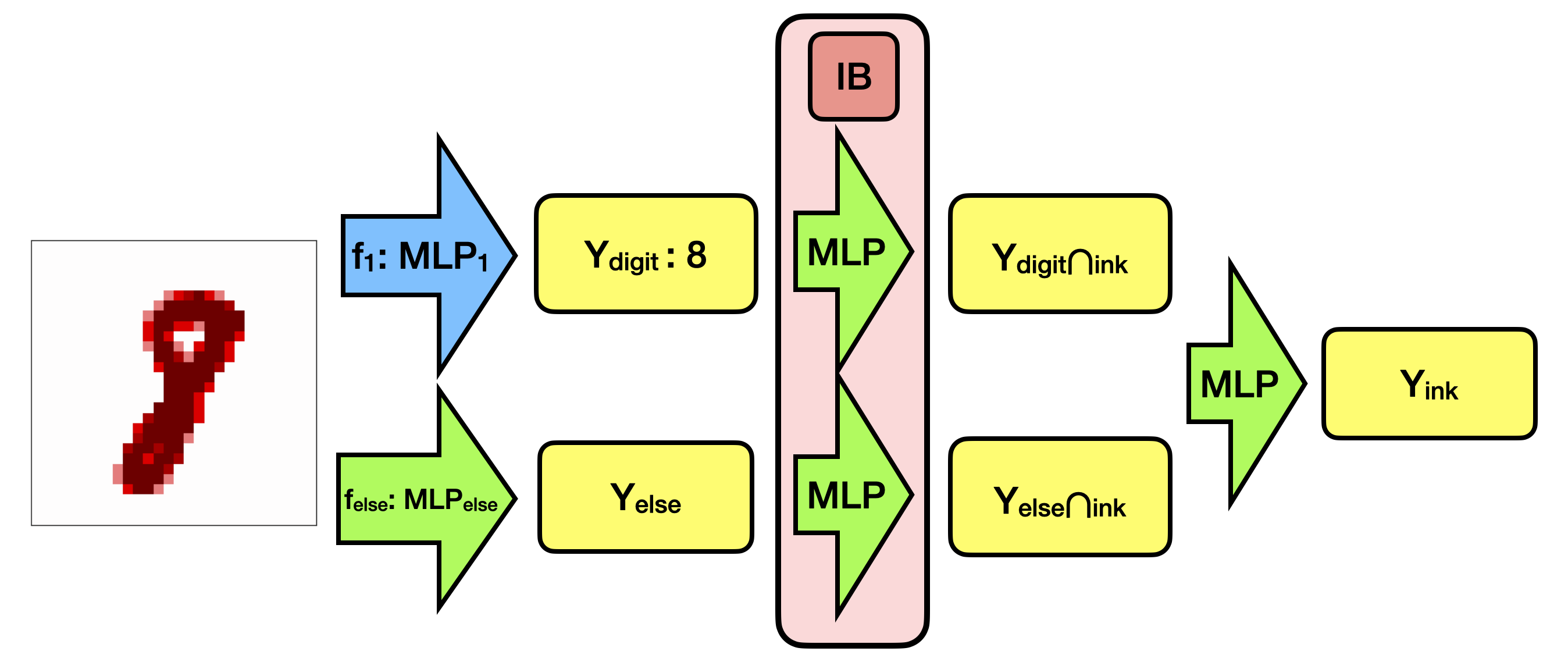}
         \caption{Learning and interpretation of later tasks (such as the ink task).}
         \label{inktask}
     \end{subfigure}
        \caption{The computational graph for MNIST dataset tasks.}
        \label{mnistgraph}
\end{figure}

\textbf{Step 1, the property learning and information partition splitting step (Fig. \ref{digitMLP}).} To solve task 1, we train a multi-layer perceptron (MLP) $\rm MLP_1$ to predict the digit. We use cross-entropy loss between the softmax outputs and the digit labels. The digit representation $Y_{\rm digit}$ is obtained by sampling from the output distribution, which is a 10-dimension one-hot vector. Note that the noise introduced by sampling here helps remove task-irrelevant information. The network is trained with loss back-propagation with the help of Pytorch’s auto-gradient framework. AdamW with a weight decay of 0.01 is chosen as the optimizer and the learning rate is fixed at 1e-4. After training, we get 98.15\% of correct
rate on the test set.

After getting the digit information, we train $\rm MLP_{\rm else}$ to get complementary representation $Y_{\rm else}$ with the help of IB. The scaling factor of IB is $\gamma = 4e-4$. $Y_{\rm else}$ is a 64-dimension continuous vector, with each dimension sampled from a Gaussian distribution $\mathcal{N}(\mu, \sigma)$ where $\mu$ and $\sigma$ are calculated by $Y_{\rm else}$. $Y_{\rm digit}$ concatenated with $Y_{else}$ is sent to another MLP to proceed auto-encoding. Pixel-by-pixel mean-square error loss is used. 

\textbf{Step 2, second task interpretation step (Fig. \ref{inktask}).} By feeding $Y_{\rm digit}$ and $Y_{\rm else}$ to two separated IB regularized MLPs, a series of second tasks, including parity task, ink task, and matrix tasks, can be solved and interpreted. For parity task and ink task, we choose the dimension of $Y_{\rm digit \cap \rm task2}$ and $Y_{\rm else \cap \rm task2}$ to be 1 and for matrix task, we choose the dimension of $Y_{\rm digit \cap \rm task2}$ to be 1 and that of $Y_{\rm else \cap \rm task2}$ to be 4. The scaling factor of IB is chosen to be 0.005, 0.002, or 0.05 for each task, which is usually decided by scanning while balancing interpretation versus precision. The experiment is conducted on a single Tesla V100S GPU.

\subsubsection{CLEVR Case}
\label{sec:8-1-2}

\begin{figure}[t]
     \centering
     \begin{subfigure}[t]{0.8\textwidth}
         \centering
         \includegraphics[width=\textwidth]{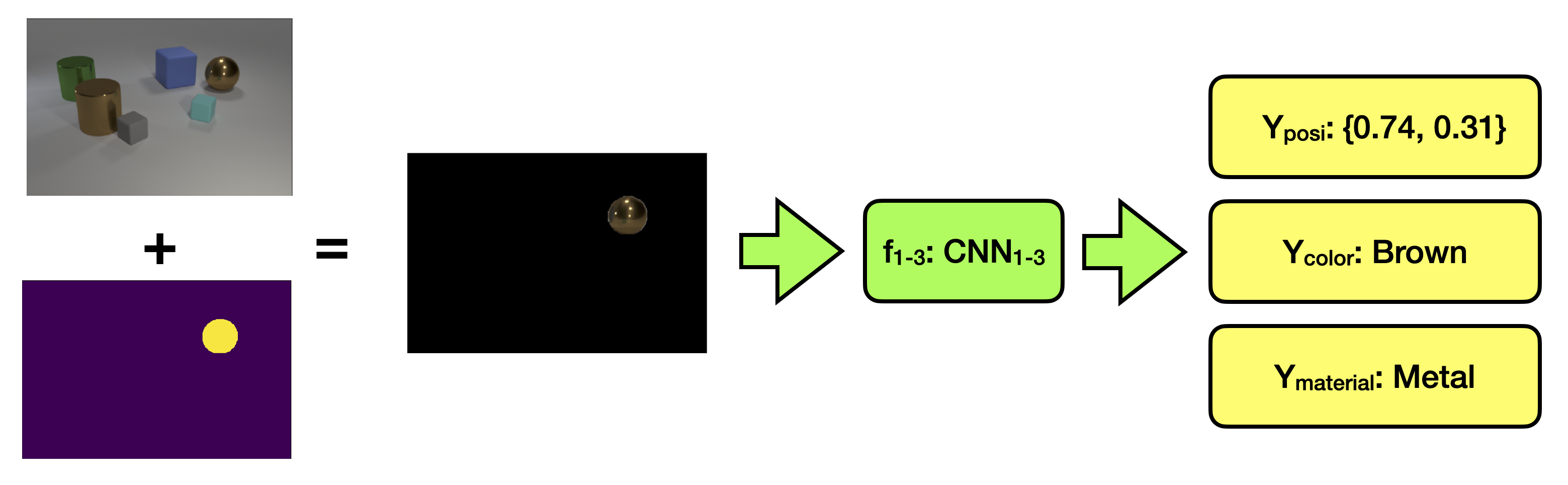}
         \caption{Property supervised learning by $\mathrm{CNN}_{1-3}$.}
         \label{propCNN}
     \end{subfigure}
     \begin{subfigure}[t]{0.8\textwidth}
         \centering
         \includegraphics[width=\textwidth]{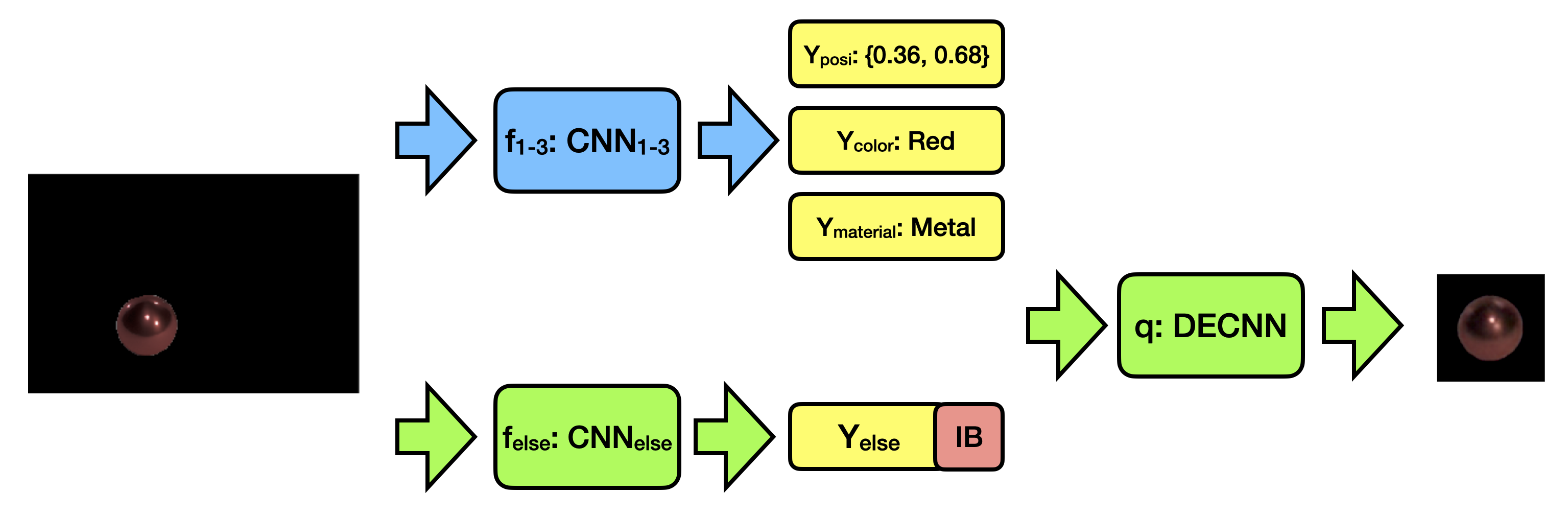}
         \caption{Information map splitting step via auto-encoding.}
         \label{Autoencode}
     \end{subfigure}
     \begin{subfigure}[t]{0.8\textwidth}
         \centering
         \includegraphics[width=\textwidth]{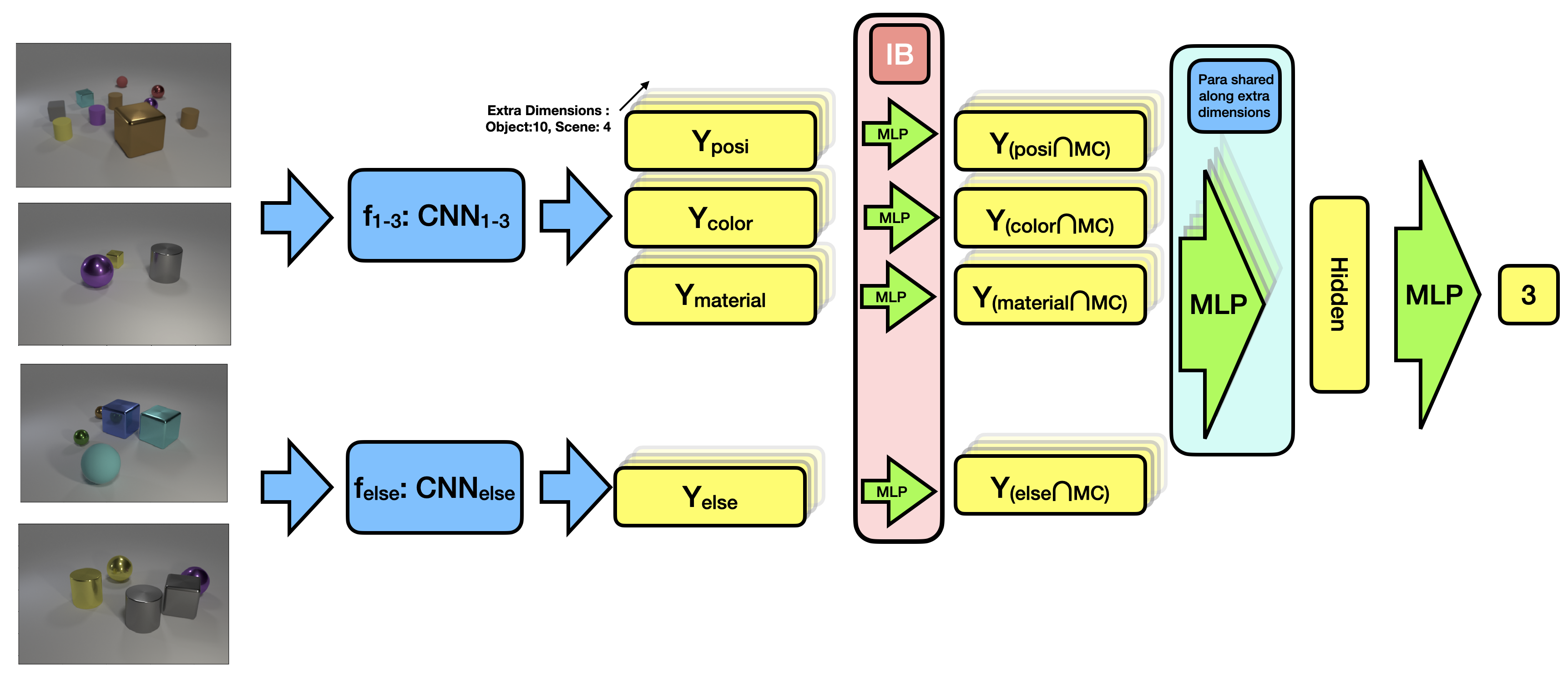}
         \caption{MC task learning and interpretation.}
         \label{interp}
     \end{subfigure}
        \caption{The computational graph for simplified CLEVR dataset interpretation.}
        \label{compgraph}
\end{figure}

\textbf{Step 1, the property learning (Fig. \ref{propCNN}) and information partition splitting step (Fig. \ref{Autoencode}).} Before the multiple-choice task, we pre-train the model to tell different objects apart and train it to recognize certain properties. To tell objects apart, the first step we need to do is image segmentation. We implement image segmentation with Mask R-CNN \citep{he2017mask} via fine-tuning a pre-trained Pytorch \citep{paszke2019pytorch} Mask R-CNN on a mini-CLEVR dataset where 4000 pictures are annotated with masks \citep{yi2018neural}. The CLEVR dataset contains the ground truth of the rendered position, color, shape, size, and material of each object. Since our framework is progressive, we assume that the model will learn about the position, color, and material of the objects first. Following the segmentation process, the masked image of an object is sent to a Res-Net \citep{he2016deep} that learns position, color, and material with supervision. The network output for object position is a continuous number. The root-mean-square error loss is used to quantify the position prediction. We add to the network estimated position Gaussian noise with a standard deviation equivalent to 0.2\% of the image size to eliminate the small amount of position-independent information. In contrast, color and material labels are categorical. The corresponding softmax outputs of the network represent the underlying probability of the one-hot categorical representation. The cross-entropy loss between the softmax outputs and the categorical labels is used for color and material prediction. Then, the internal representation for color and material is obtained by sampling from the softmax outputs. Again, this sampling step helps to eliminate task-irrelevant information encoded in the network outputs. The network is trained with loss back-propagation with the help of Pytorch's auto-gradient framework. Adam is chosen as the optimizer and the learning rate is fixed at 1e-4.

After getting the position, color, and material-related information partitions, IB-induced auto-encoding can be used as the next step to extract the complementary representation $Y_{\rm else}$. $f_{1-3} = \rm{CNN}_{1-3}$ trained in the last step is fixed in this step, providing information about the position, color, and material. CNN represents a convolutional neural network \citep{lecun1989backpropagation}. $f_{\rm else,\theta} = \rm{CNN}_{\rm else}$ is trained to encode information other than that previously learned with the help of an IB. The scaling factor of IB is $\gamma=8{\rm e}-3$. Information coming from both $\rm{CNN}_{1-3}$ and $\rm{CNN}_{\rm else}$ are combined and fed into a de-convolutional neural network (DECNN) \citep{zeiler2010deconvolutional} to do self-prediction. Pixel-by-pixel mean-square error loss is used for self-prediction. In practice, we found out that reconstructing a cropped region where the target object is in the center instead of the original masked picture with a large background area significantly improves the auto-encoding precision. 

After this step, we obtain an internal representation of $Y=\{Y_{\rm posi},Y_{\rm color}, Y_{\rm material}, \\ Y_{\rm else}\},$ where each partition represents information for the position, color, material, and other un-conceptualized properties about the input. $Y_{\rm posi}$ is a 2D float number between 0 and 1 representing normalized X and Y positions. $Y_{\rm color}$ is a one-hot vector with thr length 8, representing 8 different colors. $Y_{\rm material}$ is a one-hot vector with length 2, representing 2 different kinds of materials. $Y_{\rm else}$ is the Bernoulli distribution with dimension size 64. This internal representation will be used to solve the multiple-choice task described below and serve as the basis for the task interpretation.

\textbf{Step 2, the multiple-choice task interpretation step (Fig. \ref{interp}).} The task is a multiple-choice task regarding the categorized position (right, middle, and left), color, material, shape, and size. After choosing a certain criterion, for example: “chose the picture with a green ball,” the model is asked to pick the picture with a green ball from four candidates. To keep the task simple, we do not provide the explicit question description in natural language, and instead, we present the model with a lot of multiple-choice task examples and choice answers, and the model is supposed to find out how to solve the task without having access to the question context in natural language. In our task example generation system, only one of four images would satisfy the requirement. The pictures are randomly shuffled so that the probability of the correct choice is equal to 1/4 for all four options.

The $\rm CNN_{1-3}$ trained in previous Step 1, which receives a masked image of an object and outputs information partition $Y=\{Y_{\rm posi},Y_{\rm color},Y_{\rm material},Y_{\rm else}\},$ per object, will be fixed in this step. Each information partition is then fed into an IB regularized MLP separately, followed by a feed-forward style neural network. The scaling factor of IB is chosen to be $\gamma = 0.04$ for this step. The feed-forward network will first do convolution with an MLP kernel over the object dimension followed by another MLP over four scenes to solve the multiple-choice problem. The experiment is conducted on a single Tesla V100 GPU.

\subsubsection{Other Implementation Details}

\textbf{Temperature Schedule for Gumbel Softmax:} Instead of fixing the temperature in Gumbel softmax at a constant low temperature, we found out that multiple scans of the temperature from high to low benefit training. We use an exponential schedule to control the Gumbel softmax temperature $\tau = \rm exp \it (-5(n \times s-\lfloor n \times s \rfloor))$ where $n$ is the total number of scans, and $s$ is the training schedule that starts with 0 and ends with 1. $\lfloor... \rfloor$ is the floor operator.

\textbf{IB scaling factor $\gamma$ Schedule:} The work of \cite{shwartz2017opening} claims that training a deep network usually consists of two phases: training the label-fitting phase and the information-compression phase. Inspired by this work, we try adding a warm-up training phase where IB scaling factor $\gamma$ is set to zero and use learning rate 1e-4 to train the network. After that, the IB scaling factor is set back to normal, and information starts to get compressed. This strategy especially works well with the multiple-choice task, where we encountered some cases where the loss function never goes down if we start training with a non-zero IB scaling factor.

\textbf{Mutual Information Estimation:} For mutual information (MI) estimation, in the MNIST case where the latent representation is continuous, we find that IB would decouple each dimension, so we calculate the MI of each dimension separately and add them up. In the CLEVR case where the latent representations are discrete, we can directly use the definition to calculate MI.

\subsection{Network Implementation Detail}
\label{sec:8-2}

\subsubsection{MNIST Case}

In this section, we describe the detailed implementation of neural networks solving the MNIST tasks. We use mainly MLP. $\rm{MLP}_1$, which does digit recognition, $\rm{MLP}_{else}$, which does auto-encoding, and $\rm{MLP}_{q}$, which decodes the image, are all MLPs with 5 hidden layers with dimension 800 for each layer. The latent code vector produced by $\rm{MLP}_{else}$ has a dimension of 64. All the other auxiliary MLPs share the same structure with 5 hidden layers with dimension 128 for each layer. We also noticed that providing the digit information predicted by $\rm{MLP}_1$ to $\rm{MLP}_{else}$ can help remove digit information, so we concatenate a linear mapping of the predicted digit together with the image as the input to $\rm{MLP}_{else}$.

\subsubsection{CLEVR Case}

\begin{figure}[ht]
\centering
\includegraphics[width=0.98\linewidth]{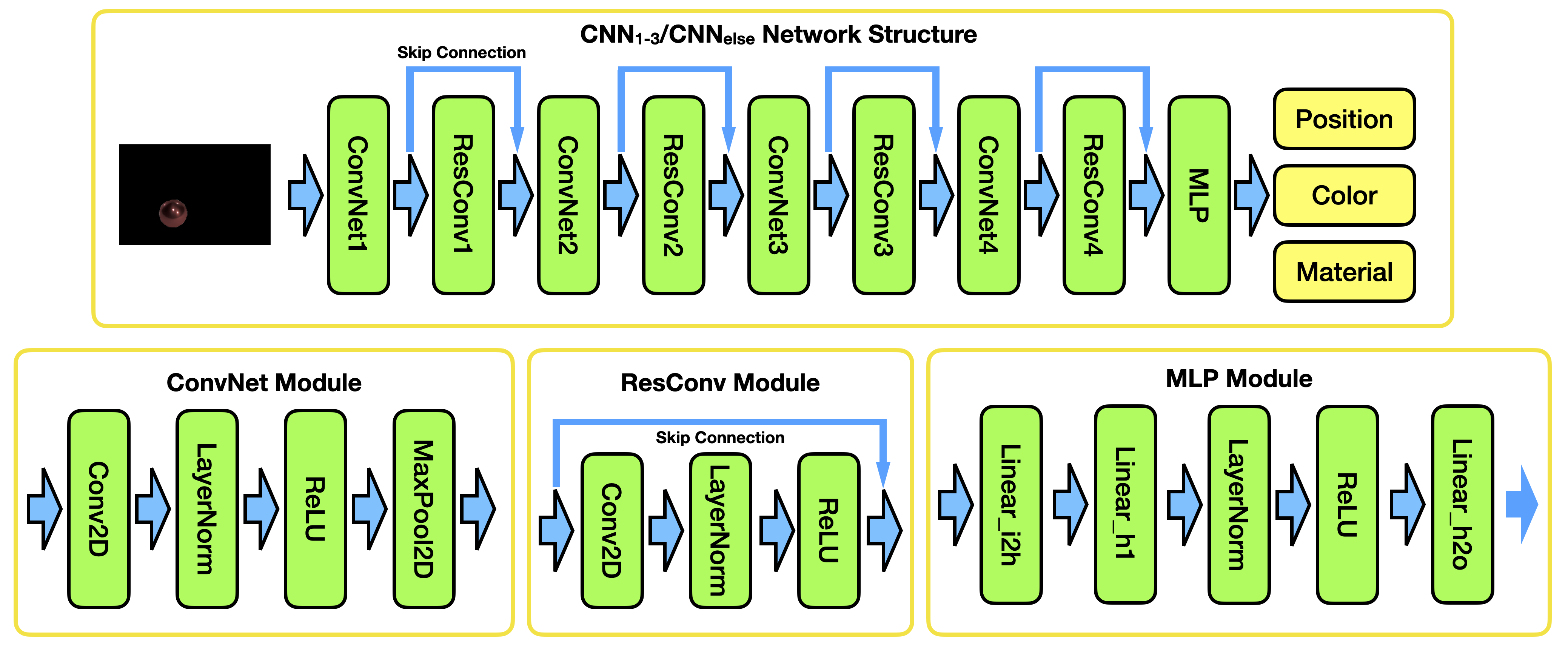}
\caption{Network topology for $\rm{CNN}_{1-3}$ and $\rm{CNN}_{else}$}
\label{cnn1net}
\end{figure}

\begin{figure}[ht]
\centering
\includegraphics[width=0.98\linewidth]{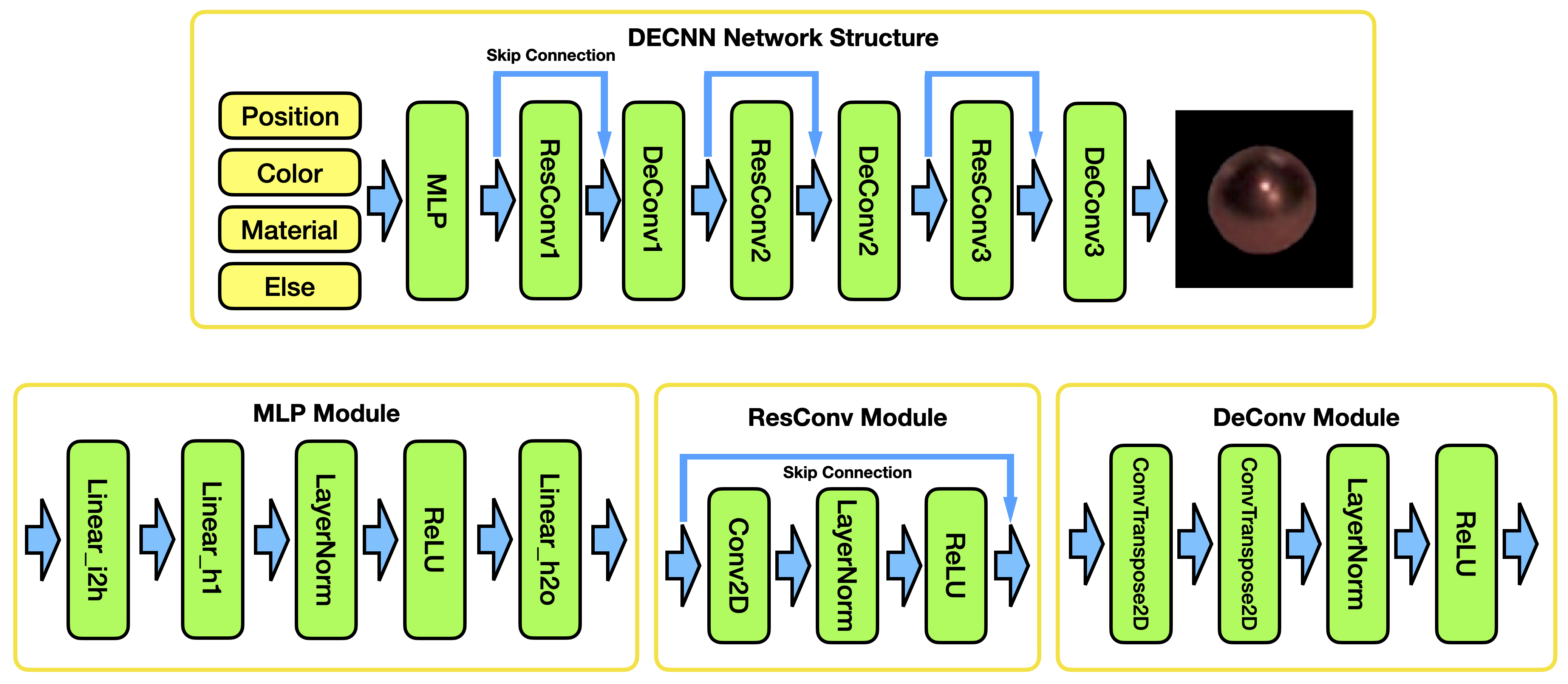}
\caption{Network topology for DECNN}
\label{decnnnet}
\end{figure}

In this section, we describe the detailed implementation of neural networks solving the CLEVR tasks. Fig \ref{cnn1net} shows the topology of the convolutional network for $\rm{CNN}_{1-3}$ and $\rm{CNN}_{else}$ of the auto-encoding step. The model parameters are listed in Table \ref{cnn1-table}. Fig \ref{decnnnet} shows the topology of the deconvolutional neural network DECNN of the auto-encoding step. Detailed model shape parameters are listed in Table \ref{decnn-table}. MLP modules of the interpretation step map the target dimension of the input tensors from a certain input size to a certain output size with several hidden layers. Each hidden layer is followed by a layer-wise normalization and a ReLU activation. Detailed shape parameters for each MLP module are shown in Table \ref{mlp-table}. (Abbreviations, CNN: convolutional neural network. DECNN: deconvolutional neural network. MLP: multi-layer perceptron. Conv2D: 2D convolutional layer. ResConv: convolutional layer with skip connection. LayerNorm: layer normalization. Linear: linear transformation layer. ConvTranspose2D: 2D deconvolutional layer.)

\begin{table*}
  \caption{$\rm{CNN}_{1-3}$ and $\rm{CNN}_{else}$ parameter, $\rm{CNN}_{else}$ shares the same network topology with $\rm{CNN}_{1-3}$ except for the MLP module which is shown below.}
  \label{cnn1-table}
  \centering
  \begin{tabular}{|l|ccccc|}
    \hline
    Module Name     & Size [in(x,y), out(x,y)]  &	Channel (in, out) &	Kernel &	Stride &    Padding \\
    \hline
    $\rm{CNN}_{1-3}$ parameter &&&&& \\
    \hline
    Conv2D1    &	$[(480, 320),(237, 157)]$   &	$(3, 16)$ &	7&	2&	0 \\
    MaxPool2D1 &    $[(237, 157),(118, 78)]$    &	$(16, 16)$&	2&	2&	0 \\
    ResConv1   &	$[(118, 78),(118, 78)]$     &	$(16, 16)$&	5&	1&	2 \\
    Conv2D2    &	$[(118, 78),(114, 74)]$     &	$(16, 32)$&	5&	1&	0 \\
    MaxPool2D2 &    $[(114, 74),(57, 37)]$      &	$(32, 32)$&	2&	2&	0 \\
    ResConv2   &	$[(57, 37),(57, 37)]$       &	$(32, 32)$&	5&	1&	2 \\
    Conv2D3    &	$[(57, 37),(53, 33)]$       &	$(32, 32)$&	5&	1&	0 \\
    MaxPool2D3 &    $[(53, 33),(26, 16)]$       &	$(32, 32)$&	2&	2&	0 \\
    ResConv3   &	$[(26, 16),(26, 16)]$       &	$(32, 32)$&	5&	1&	2 \\
    Conv2D4    &	$[(26, 16),(22, 12)]$       &	$(32, 32)$&	5&	1&	0 \\
    MaxPool2D4 &    $[(22, 12),(11, 6)]$        &	$(32, 32)$&	2&	2&	0 \\
    ResConv4   &	$[(11, 6),(11, 6)]$         &	$(32, 32)$&	5&	1&	2 \\
    Linear\_i2h   &	$(2112, 128)$               &	-      &	-&	-&	- \\
    Linear\_h1    &	$(128, 64)$                 &	-      &	-&	-&	- \\
    Linear\_h2o   &	$(64, 12)$                  &	-      &	-&	-&	- \\
    \hline
     $\rm{CNN}_{else}$ parameter &&&&& \\
    \hline
    Linear\_i2h   &	$(2112, 512)$                &	-      &	-&	-&	- \\
    Linear\_h1    &	$(512, 256)$                 &	-      &	-&	-&	- \\
    Linear\_h2o   &	$(256, 128)$                 &	-      &	-&	-&	- \\
    \hline
  \end{tabular}
\end{table*}

\begin{table*}
  \caption{DECNN parameter, note that the images which DECNN works with has equal x and y sizes.}
  \label{decnn-table}
  \centering
  \begin{tabular}{|l|ccccc|}
    \hline
    Module Name     & Size (in, out)                 &	Channel (in, out) &	Kernel &	Stride &    Padding \\
    \hline
    Linear\_i2h   &	$(140, 256)$    &	-      &	-&	-&	- \\
    Linear\_h1    &	$(256, 512)$    &	-      &	-&	-&	- \\
    Linear\_h2o   &	$(512, 1152)$   &	-      &	-&	-&	- \\
    ResConv1            &	$(6, 6)$      &	$(32, 32)$&	5&	1&	2 \\
    ConvTranspose2D1\_1 &	$(6, 12)$     &	$(32, 32)$&	2&	2&	0 \\
    ConvTranspose2D1\_2 &	$(12, 16)$    &	$(32, 32)$&	5&	1&	0 \\
    ResConv2            &	$(16, 16)$    &	$(32, 32)$&	5&	1&	2 \\
    ConvTranspose2D2\_1 &	$(16, 32)$    &	$(32, 16)$&	2&	2&	0 \\
    ConvTranspose2D2\_2 &	$(32, 36)$    &	$(16, 16)$&	5&	1&	0 \\
    ResConv3            &	$(36, 36)$    &	$(16, 16)$&	5&	1&	2 \\
    ConvTranspose2D3\_1 &	$(36, 72)$    &	$(16, 16)$&	2&	2&	0 \\
    ConvTranspose2D3\_2 &	$(72, 147)$   &	$(16, 3)$ &	5&	2&	0 \\
    \hline
  \end{tabular}
\end{table*}

\begin{table*}
  \caption{MLP module parameters for multiple-choice tasks, x size equals y size. b stands for batch size.}
  \label{mlp-table}
  \centering
  \begin{tabular}{|l|cccc|}
    \hline
    Module Name & Input data shape &  Input size    &	Output size &	Hidden size \\
    \hline
    MLP:$Y_{\rm posi}$      & (b, 4, 10, 2)  & 2  & 8  &	(16, 16) \\
    MLP:$Y_{\rm color}$     & (b, 4, 10, 8)  & 8  & 8  &	(16, 16) \\
    MLP:$Y_{\rm material}$  & (b, 4, 10, 8)  & 2  & 8  &	(16, 16) \\
    MLP:$Y_{\rm else}$      & (b, 4, 10, 64) & 64 & 16 &	(32, 32) \\
    MLP:$Y$ to hidden         & (b, 4, 10, 40) & 40 & 1  &	(32, 16, 8) \\
    MLP:hidden to out       & (b, 4, 10)     & 10 & 1  &	(5) \\
    \hline
  \end{tabular}
\end{table*}

\subsection{Reparameterization}
\label{sec:8-3}

The internal representation of a VIB $Y$ can be reparameterized into a vector of either continuous or discrete variables. 

\subsubsection{Continuous Representation}

One standard way to reparameterize $Y$ is to assume multi-dimensional Gaussian distribution. The $d$-th element of $Y$ is given by:
\begin{equation} \label{gumbel}
[Y]_d = \mathcal{N}([f_\mu(X)]_d, [f_\sigma(X)]_d)
\end{equation} 
where $\mathcal{N}$ is Gaussian distribution. $[f_\mu(X)]_d$ is the $d$-th element of a vector calculated from $X$ by the neural network $f_\mu$ representing the mean, and $[f_\sigma(X)]_d$ representing the variance. And we usually choose $r(Y_d) = \mathcal{N}(0, 1)$, or unit Gaussian with 0 mean and unit variance as the prior distribution. Then, the KL-divergence between $p(Y|X)$ and $r(Y)$ can be analytically calculated.

\subsubsection{Discrete Representation}

We can assume $Y$ to be a vector of binary elements with multi-dimensional Bernoulli distributions. One merit of multi-dimensional Bernoulli distribution is that we can regularize the prior distribution's entropy to induce sparsity and make the following analysis easier, which is usually not the case for continuous prior such as multi-dimensional Gaussian. The merit of entropy regularization is more thoroughly discussed in the work of deterministic information bottleneck of \cite{strouse2017deterministic}. Specifically, we use the Gumbel-Softmax reparameterization trick \citep{jang2016categorical} to draw samples from the multi-dimensional Bernoulli distribution without blocking gradient information. The $d$-th element of $Y$ is given by:
\begin{equation} \label{gumbel}
[Y]_d = {\rm Sigmoid}\left[\left(\log \frac{[f_{\theta}(X)]_d}{1-[f_{\theta}(X)]_d} + \Delta g_d\right)/\tau\right]
\end{equation} 
where ${\rm Sigmoid}[x]=1/(1+e^{-x})$ is the Sigmoid function, $\tau$ is a temperature parameter, and $\Delta g_d$ is the difference of two Gumbel variables, while each of them is generated by $-\log\log(1/\sigma)$ with $\sigma$ being a sample from uniform distribution Uniform(0,1) between 0 and 1. Theoretically, we consider the zero temperature limit $\tau\to 0$ of Eq.~\ref{gumbel} to obtain the binary representation, namely,  $[Y]_d=1$ with probability $[f_{{\rm else},\theta}(X)]_d$ and $[Y]_d=0$ otherwise. In practice, however, we observed that scanning the temperature from high to low multiple times helps the network to converge. $r(Y) = \prod_d (\hat{r}_d)^{[Y]_d}(1-\hat{r}_d)^{1-[Y]_d} $ is the Bernoulli prior distribution for $p$ with parameters $\{\hat{r}_d\}$, which should be optimized. Note that the KL-divergence between $p(Y|X)$ and $r(Y)$ can be analytically calculated.

\subsection{Splitted Information Map Combinatory Test}
\label{sec:8-4}

\begin{figure}[ht]
\centering
\begin{subfigure}[t]{0.98\linewidth}
\includegraphics[width=0.98\linewidth]{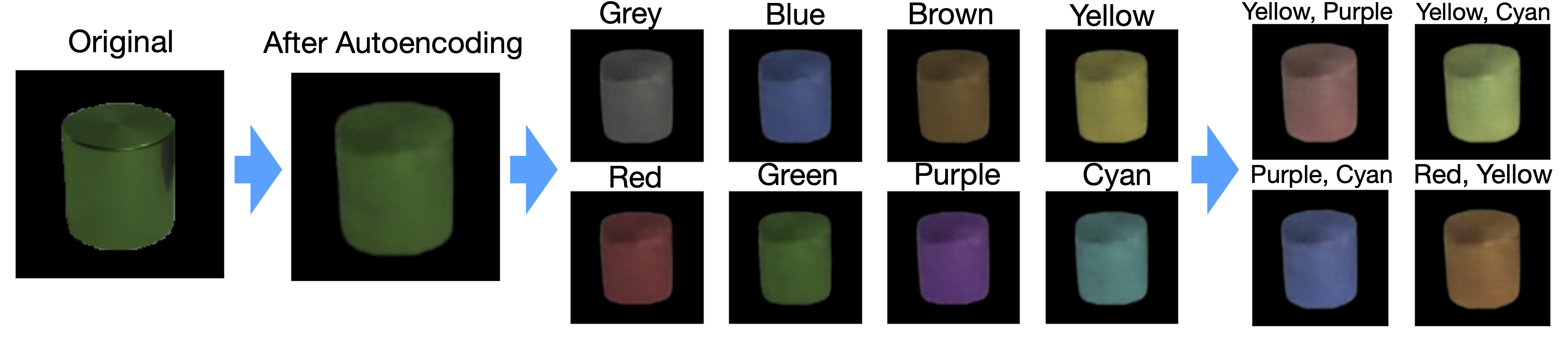}
\caption{Color combinatory test. From left-to-right shows the original figure, auto-encoded figure, colors changed, and colors mixed.}
\label{colorComb}
\end{subfigure}
\begin{subfigure}[t]{0.98\linewidth}
\centering
\includegraphics[width=0.75\linewidth]{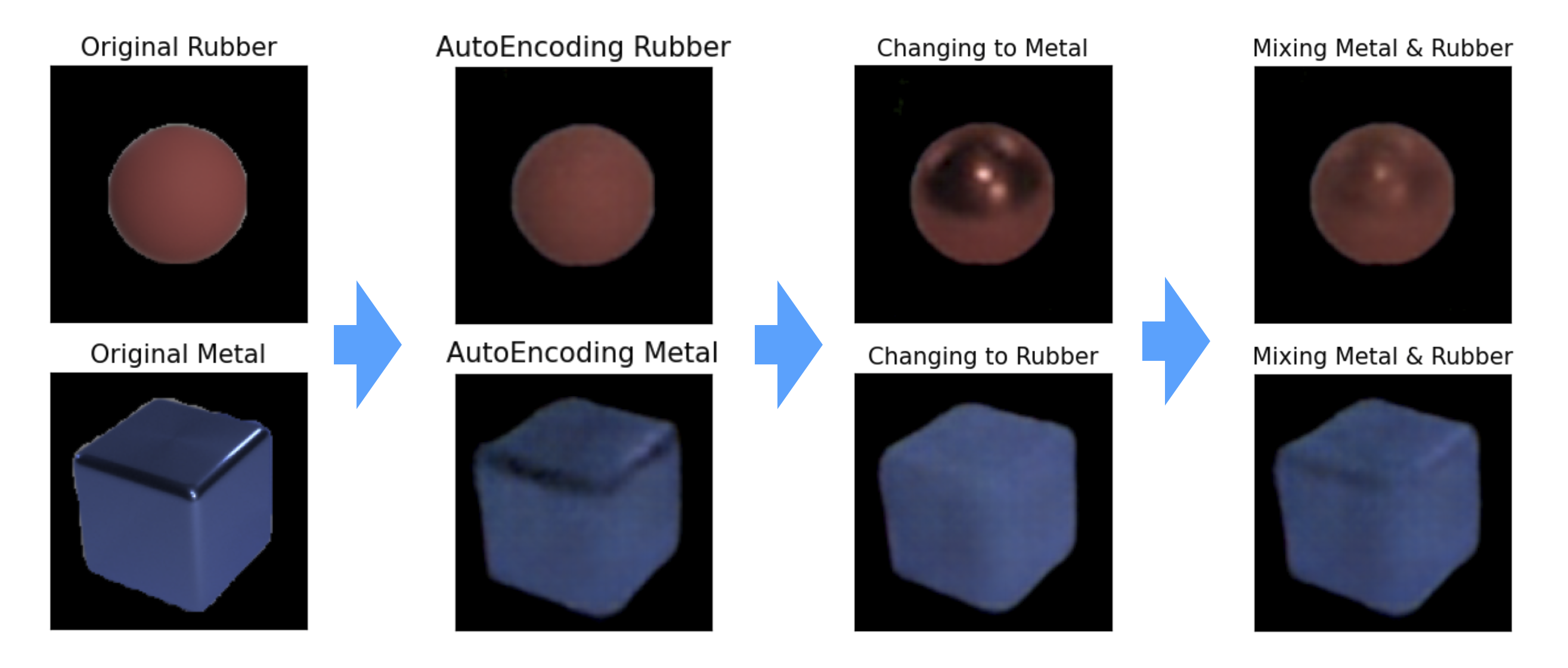}
\caption{Material combinatory test. From left-to-right shows the original figure, auto-encoded figure, materials changed, and materials mixed.}
\label{MatComb}
\end{subfigure}
\caption{Splitted information map combinatory test.}
\end{figure}

The auto-encoding step splits input representation $Y$ into $Y_1$, which is needed to solve task 1, and $Y_{\rm else}$, which ideally satisfies $I(Y_1;Y_{\rm else})=0$. This means $Y_1$ and $Y_{\rm else}$ should be independent. To check the quality of this auto-encoding step, except for calculating mutual information and scanning latent variables as shown in the MNIST example, another way is to test the combinatory property of $Y_1$ and $Y_{\rm else}$ visually. We demonstrate it using the CLEVR dataset. For example, Fig. \ref{colorComb} shows when $Y_1$ classifies object color. Since $Y_1$ is a one-hot vector with eight possible color choices, it is straightforward to change the one-hot vector, combine it with $Y_{\rm else}$ and generate new auto-encoded pictures. The result shows that, after changing the color representation, the output image color also changes accordingly, while other visual properties are fixed. One can even try to mix the color component by mixing the color representation vectors and generating an object with new color outside the dataset. The same experiment can be done when $Y_1$ classifies object material. One can easily switch the object material between rubber and metal without changing other visual properties. This experiment confirms that the information splitting method can generate combinatory information partitions.

\subsection{Hyper Parameter Scan}
\label{sec::gamma}

The scaling factor $\gamma$ is a hyper-parameter balancing inclusion of more information for better performance and exclusion of information for simplicity. To study the effect of $\gamma$, we perform $\gamma$ scanning in this section. We study the CLEVR dataset as an example.

\subsubsection{$\gamma$ in Auto-encoding step}

\begin{figure}[ht]
\centering
\includegraphics[width=0.98\linewidth]{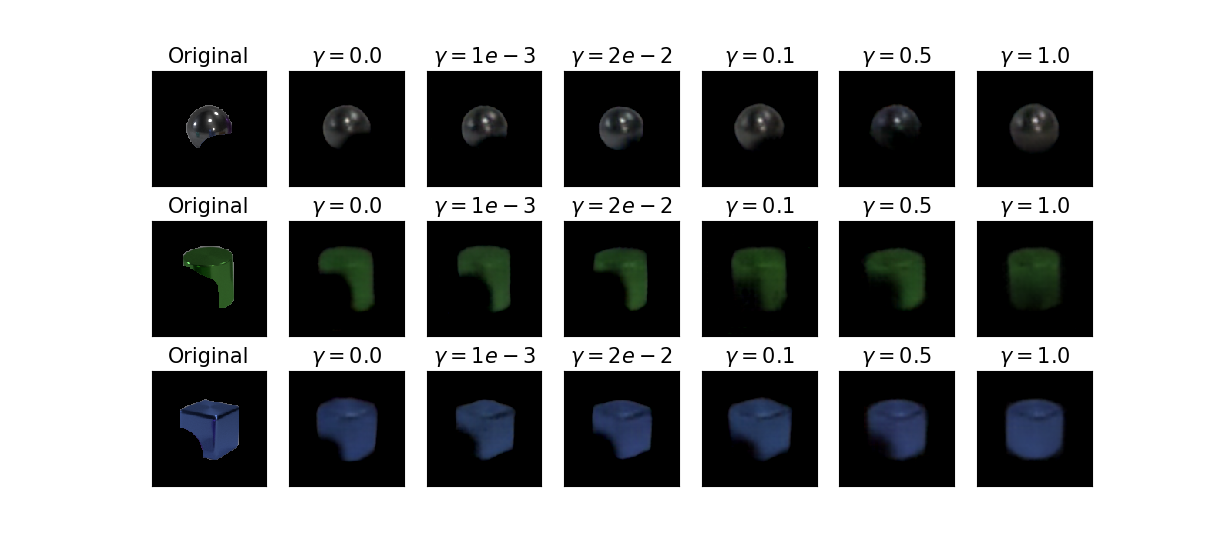}
\caption{The reconstructed picture with different $\gamma$}
\label{autoscan_pic}
\end{figure}

The first $\gamma$ subject to scan is the $\gamma$ in Eq. 1 of the main manuscript. We pick $\gamma$ from \{ 0.0, 1e-3, 8e-3, 2e-2, 0.1, 0.5, 1.0 \}. Note that since the auto-encoding step uses RMS error as the training loss, which is not an information metric, the absolute value of $\gamma$ has no meaning. Fig. \ref{autoscan_pic} shows the result of reconstructed objects with different $\gamma$. It shows that generally speaking, higher $\gamma$ leads to a more generic picture. One interesting thing to notice is that when $\gamma$ becomes 1.0, the model decides to fill in the covered parts of the image. 

\begin{figure}[ht]
\centering
\includegraphics[width=0.98\linewidth]{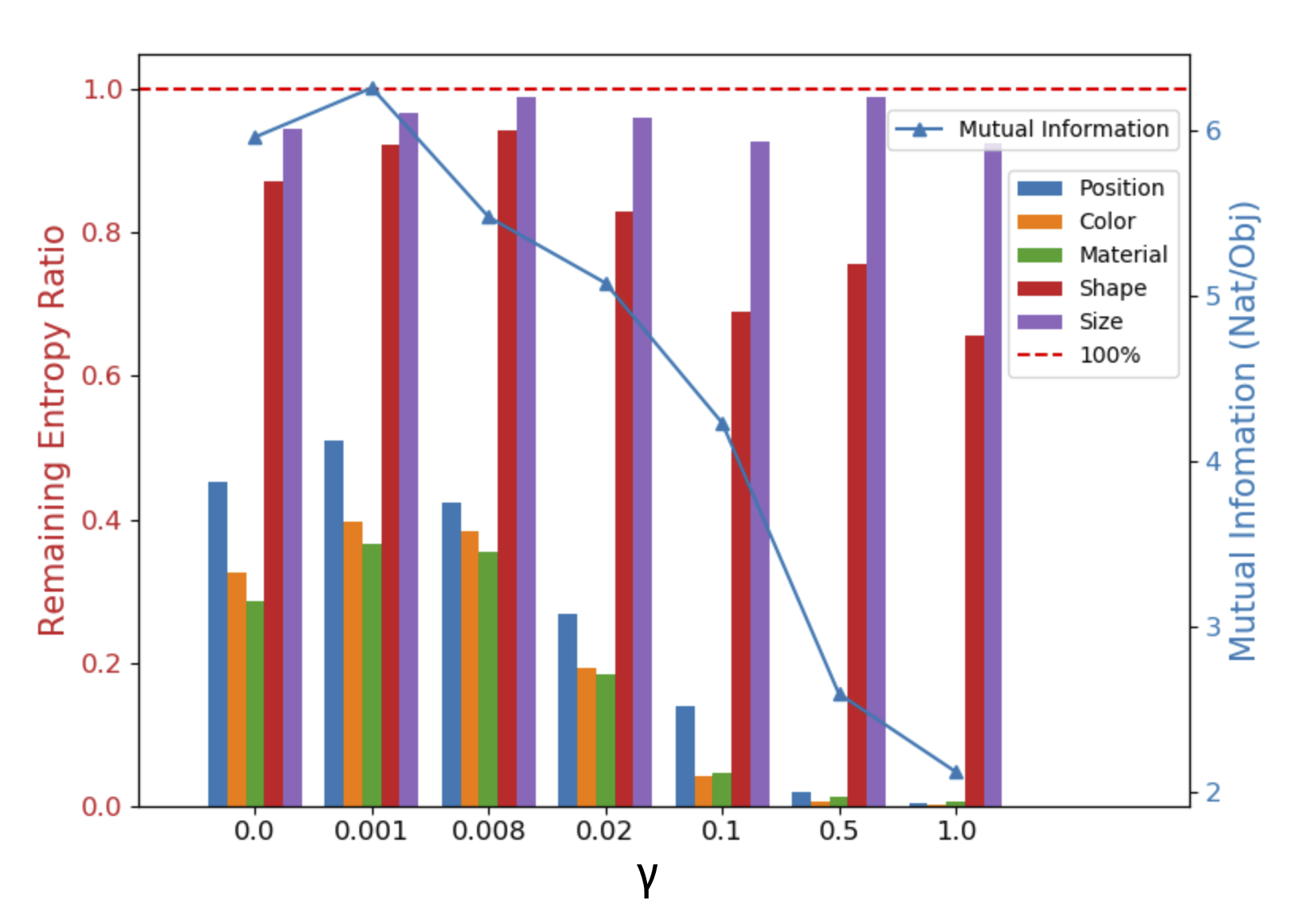}
\caption{Information metric with different $\gamma$ in the auto-encoding step. The bar plot (corresponding to the left axis) shows the ratio of information remained in the compressed representation $I(Y_n;Y_{\rm{else}})/H(Y_n), n \in \{\rm{position, color, material, shape, size}\}$. The triangle plot shows total mutual information $I(X; Y_{\rm{else}})$ measured in Nat per object.}
\label{autoscan_info}
\end{figure}

Fig \ref{autoscan_info} shows the information metric about $Y_{\rm{else}}$ with different $\gamma$. With $\gamma$ increasing, total mutual information $I(X; Y_{\rm{else}})$ goes down. Mutual information about the position, color, and material, which is supposed to be compressed out, decreases since they are already encoded by $Y_{1-3}$. However, as a side effect, mutual information about the unknown property (shape and size), which should be kept, slightly goes down. Feasible $\gamma$ ranges from 0 to 0.02. Note that choosing $\gamma$ to be 0 doesn't mean the information bottleneck is not needed, since discretization itself is already a very strong bottleneck. We choose $\gamma=8e-3$ with higher mutual information about shape and size, granting better downstream tasks performance.

\subsubsection{$\gamma$ in Multiple-choice Task}

\begin{figure}[ht]
\centering
\includegraphics[width=0.98\linewidth]{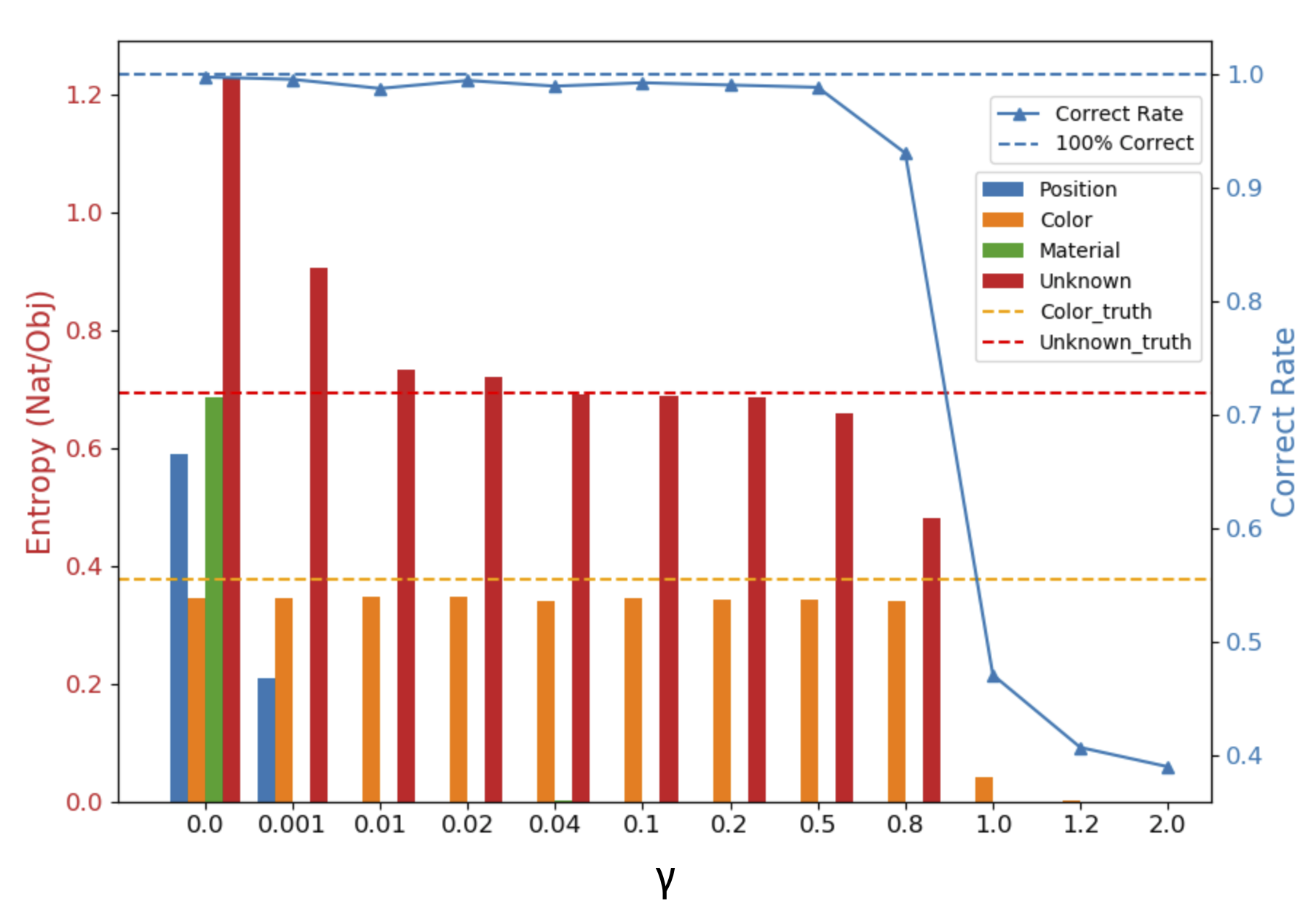}
\caption{Information metric and correct rate of the multiple-choice task ``exist small yellow object" v.s. different $\gamma$. The bar plot shows mutual information from different information partitions (corresponding to the left axis). The orange dash line shows ground truth mutual information for the color partition. The red dash line shows ground truth mutual information for the unknown partition. Note that only the color and unknown partitions are required to solve this task. The blue triangle shows the task correct rate (corresponding to the right axis) with the blue dash line being 100\% as a reference.}
\label{mcgammascan}
\end{figure}

Fig. \ref{mcgammascan} shows the scan of $\gamma$ of the multiple-choice step in Eq. 4 of the main manuscript. Among the six tasks, we choose ``exist a small yellow object" as the example. We pick $\gamma$ from \{ 0.0, 1e-3, 1e-2, 2e-2, 4e-2, 0.1, 0.2, 0.5, 0.8, 1.0, 1.2, 2.0 \}. As shown by the result, the correct rate doesn't change much when $\gamma$ is smaller than 0.5 and drops fast afterward. If $\gamma$ is too small, for example, 0 as an extreme case, information is coming from all sub-partitions, including the unnecessary position and material partitions for solving this task. Hence, too small $\gamma$ hampers the interpretation due to the contamination. Feasible $\gamma$ with both high correct rate and high-quality interpretation ranges from 0.01 to 0.2.  

\subsection{Failure Case Analysis}

Table 1 of the main manuscript shows that there is still a significant amount of failure cases for the CLEVR task, especially the ``red cube" task and ``right side cylinder" task. Then, an interesting question to ask is, can our interpretation framework detect the reason for failure?

\begin{figure}[ht]
\centering
\includegraphics[width=0.98\linewidth]{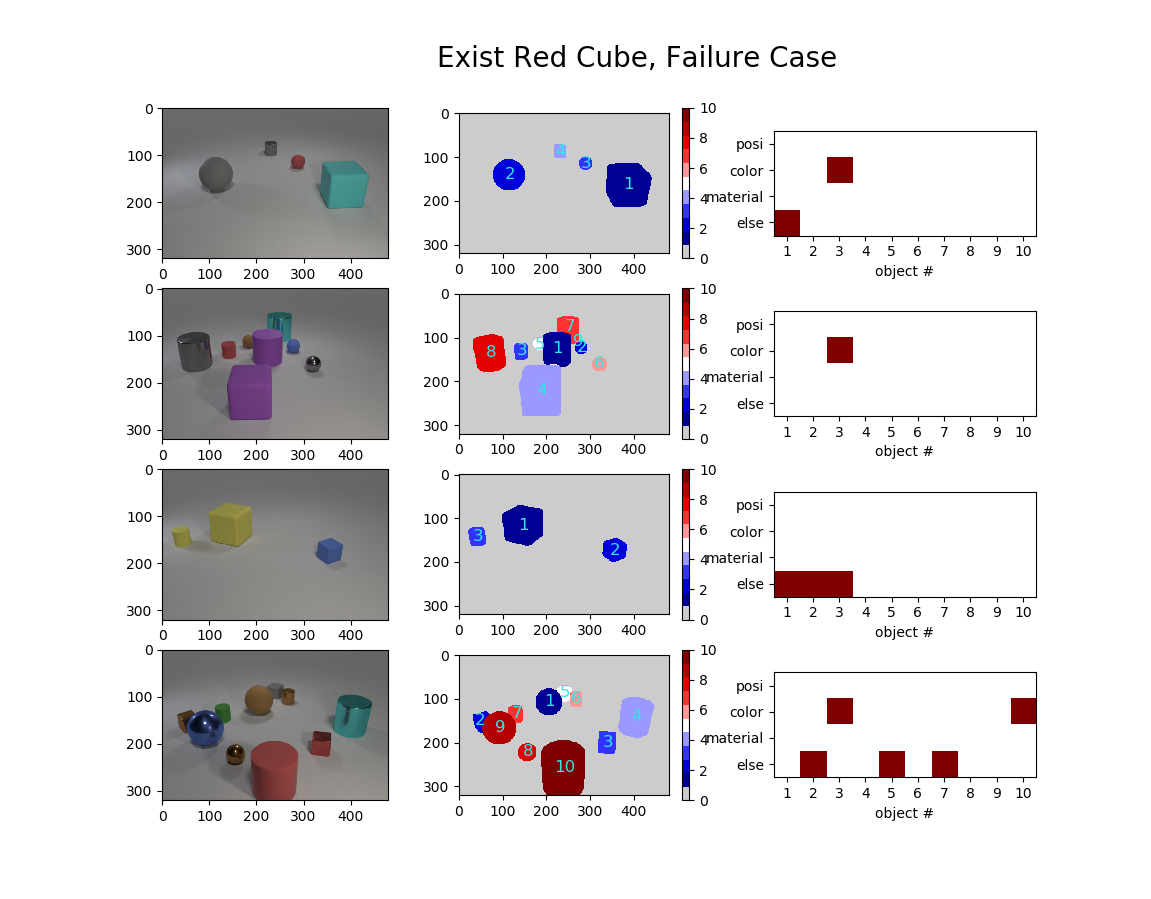}
\caption{Single-example interpretation of the question ``choose a picture with a red cube". A typical failure case example. Plot convention follows Fig. 5 of the main manuscript. Row 4 indicates a false negative due to the failure of classifying object 3 to be a cube. On the other hand, the color red can be precisely detected.}
\label{redcubefailure}
\end{figure}

The answer is yes. Fig. \ref{redcubefailure} shows the single example interpretation plot of a failed ``red cube" detection case. By checking the representation matrix, we noticed that the model is quite stable at predicting ``red" objects, but missed multiple times about ``cube" in this particular case. For example, object 4 in picture 2, which is a cube, is missed. Object 3 in picture 3, which is a cylinder, is incorrectly identified as a cube. 

\begin{figure}[ht]
\centering
\includegraphics[width=0.98\linewidth]{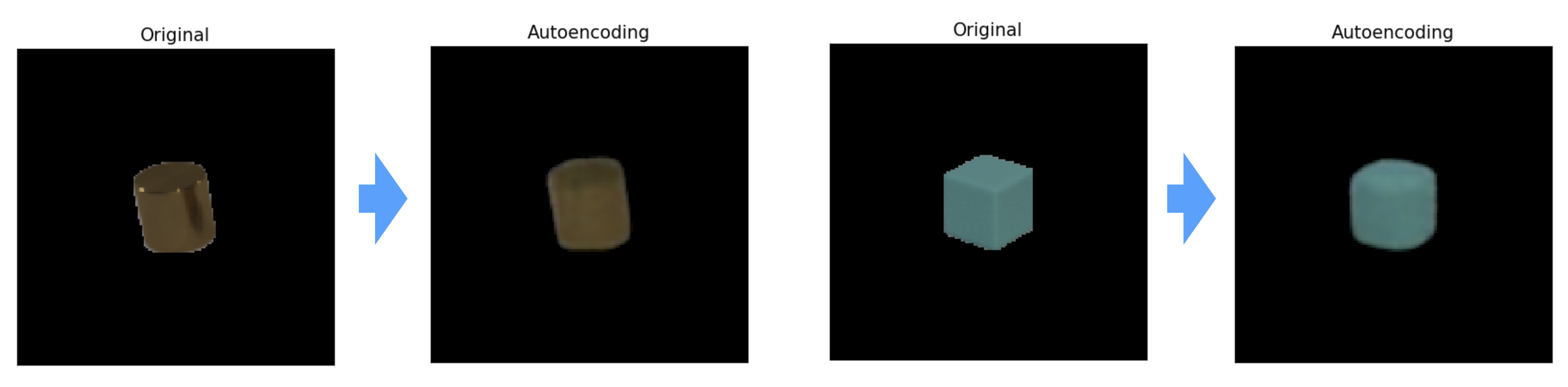}
\caption{Two examples of a cylinder and cube visual reconstruction after auto-encoding.}
\label{cylcubeauto}
\end{figure}

After checking more failure examples, we hypothesize that the model may have difficulty telling cube and cylinder apart due to the side-effect of information-bottleneck-induced auto-encoding. The hypothesis is also supported by a visual check of the auto-encoding result. Two examples are shown in Fig. \ref{cylcubeauto} where the original object lies on the left of the arrow and the reconstructed object lies on the right. The reconstructed object shows some tendency to mix cylinder and cube.

\subsection{Importance of information splitting}
\label{sec::abalation}

One key step of our proposed framework is the information partition splitting shown by Eq. 1. After splitting the input into task 1 related partition $Y_1$ and unrelated partition $Y_{\rm{else}}$, later task then has to go to $Y_1$ for task 1 related information since $Y_{\rm{else}}$ doesn't contain the necessary information. And the information accessing pattern is the key to the interpretation in our framework. 

Then, as an ablation study, it would be interesting to ask, what would happen if we remove the information partition splitting step and replace $Y_{\rm{else}}$ with simply the hidden representation $Y_{\rm{auto}}$ for auto-encoding of the input. Since now $Y_{\rm{auto}}$ also contains information of $Y_1$, there would be ambiguity for the later task to access information about task 1. We perform the same CLEVR dataset multiple-choice task described in the main manuscript with $Y_{\rm{else}}$ replaced by $Y_{\rm{auto}}$. Note that since information compression is not needed, we also remove the information bottleneck when building $Y_{\rm{auto}}$. The result is shown in Table \ref{abalation}. 

As can be seen from the result, there is no problem for the model to solve the tasks. However, if we check the information flow pattern, we can see that the model fails to relate the new task with previously learned features and get information only from the auto-encoded part. From a meta-information point of view, the information flow pattern degenerates, and the interpretation is viewed to be low quality since it cannot tell different tasks apart.

\begin{table*}[ht]
\caption{Table for multiple-choice task interpretation, without information partitioning, information unit (Nat/object)}
\label{abalation}
\begin{center}
\begin{small}
\begin{sc}
\begin{tabular}{|l|ccccc|}
\hline
Question Type&	Position &	Color &	Material &	Auto-code &	Correct rate \\
\hline
Green Metal&	$<0.001$ &	$<0.001$ &	$<0.001$ &	0.201&	99.6\% \\
Left side Rubber&	$<0.001$ &	$<0.001$ &	$<0.001$ &	0.330&	99.1\% \\
Small Yellow&	$<0.001$ &	$<0.001$ &	$<0.001$ &	0.193&	99.5\% \\
Red Cube&	$<0.001$ &	$<0.001$ &	$<0.001$ &	0.249&	99.3\% \\
Right side Cylinder&	$<0.001$ &	$<0.001$ &	$<0.001$ &	0.270&	98.2\% \\
Large Sphere&	$<0.001$ &	$<0.001$ &	$<0.001$ &	0.244 &	99.9\% \\
\hline
\end{tabular}
\end{sc}
\end{small}
\end{center}
\end{table*}

\subsection{Higher-order Information}
\label{sec::highorder}

One common question about this framework is how this can handle high-order information. Imagine the following scenario, where the correct output of a new task is independent of each partition but computed from a combination of some partitions. In this case, the task is solvable only by using high-order information. We argue that our interpretation framework can handle this situation. Recall that the contribution of the $i$th partition for solving task $n+1$ is quantified by the minimal amount of information need from this partition, $I(Y_{(i; n+1)},Y_{i})$. This quantity is distinct from the mutual information between the $i$th partition and the output, $I(Y_{(i; n+1)}; Z)$, which should be 0 in our example. In other words, the amount of information kept in $Y_{(i; n+1)}$ despite the application of the information bottleneck directly indicates the contribution of this partition for the task. 

We demonstrate our claim with a similar CLEVR dataset multiple-choice task. Everything remains the same except that we change the logical operation between two object properties from AND to XOR in the multiple-choice tasks. For example, the task ``green metal" becomes ``green XOR metal", which means ``finding the picture with the object being either green or metal but not both." In the CLEVR dataset, ``green XOR metal" is independent of either ``green" or ``metal", and the information needed to solve the task is of the second order. The interpretation result is shown in Table \ref{highorder}. We can see from the result that even though ``green XOR metal" is independent of either ``green" or ``metal", the model can still successfully relate ``green XOR metal" with ``green" and ``metal". An extra partition can still be formed when needed just as expected.

\begin{table*}[t]
\caption{Table for multiple-choice task interpretation, XOR case, information unit (Nat/object)}
\label{highorder}
\begin{center}
\begin{small}
\begin{sc}
\begin{tabular}{|l|ccccc|}
\hline
Question Type &	Position &	Color &	Material &	Unknown &	Correct rate \\
\hline
Green XOR Metal &	$<0.001$ &	0.388 &	0.575 &	$<0.001$ &	99.4\% \\
Left side XOR Rubber &	0.528 &	$<0.001$ &	0.615 &	$<0.001$ &	99.1\% \\
Small XOR Yellow &	$<0.001$ &	0.362 &	$<0.001$ &	0.574 &	99.8\% \\
Red XOR Cube &	$<0.001$ &	0.301 &	$<0.001$ &	0.581 &	97.5\% \\
Right side XOR Cylinder &	0.515 &	$<0.001$ &	$<0.001$ &	0.772 &	94.8\% \\
Large XOR Sphere &	$<0.001$ &	$<0.001$ &	$<0.001$ &	0.443 &	99.9\% \\
\hline
\end{tabular}
\end{sc}
\end{small}
\end{center}
\end{table*}

\subsection{Discussions}
\label{sec:8-9}

\paragraph{Motivation and usefulness of this framework.} Many works try to make sense of a neural network's operation at a microscopic level, while our proposal aims at a more macroscopic level interpretation. We showed that we can learn a lot by examining how the machine solves each task utilizing partitioned information relevant to previous tasks. Also different from many existing works where the goal is to interpret an already-trained network, our framework emphasizes the utilization of experience, which can be viewed as an interpretation-enhancing framework, since without information splitting, later task doesn't automatically relate to earlier tasks (see Sec. \ref{sec::abalation}). Another common concern is about the usefulness of this framework since a mutually related sequence of tasks is needed. The author agrees that, for the current stage, it is difficult to find such a dataset since the current trend is end-to-end training of a single high-level task. However, it also implies that research on related task ensembles receives non-adequate attention since we human beings naturally learn progressively. The author would like to attract more attention to this direction with this proposed framework. It will be important for future studies to learn from resources such as textbooks that are naturally organized progressively and to extend the framework onto auxiliary tasks like those used in unsupervised pretraining.

\paragraph{Relationship with partial information decomposition (PID).} Our proposed framework shares some similarities with the PID framework \citep{williams2010nonnegative} in the sense that both of them are trying to explain data by splitting the information map. However, the difference is also obvious. One obvious difference is that PID focuses on characterizing the data currently under consideration while our framework is progressive and learning-order dependent (see above), focusing on characterizing future data. Importantly, the number of information maps grows combinatorially with the number of neurons in the PID framework, while in our framework, the number of information splits grows linearly with tasks thanks to the training framework that actively simplifies the information map by introducing independence. Note that even though our framework tends to remove redundancy, synergy can still exist between current information splits and future tasks (See Sec. \ref{sec::highorder}).

\paragraph{Relationship between interpretation quality and performance.} As we explored in Sec. \ref{sec::gamma}, there exists a trade-off between interpretation quality and model performance in our proposed framework. In practice, we noticed that lowering IB regularization may result in better task performance at the cost of using more information than necessary. This leads to more redundant information partitions and an overestimation of task-relevant information. However, exchanging model performance for better understanding is not just an issue particular to our framework but is something universal. This is also the case for scientific theorems. An appropriate level of abstraction is crucial for intuitiveness and interpretability. Thus, a good balance between interpretation and performance may be the key to upgrading a model into insightful knowledge.

\paragraph{Intuition behind auxiliary auto-encoding.} We mentioned in Sec. \ref{sec:MNISTBench} that allowing $Y_{\rm else}$ to be re-trained by adding an auxiliary auto-encoding task when learning task 2 would boost the task performance. Here, we discuss the intuition why we sometimes need this auxiliary auto-encoding task. Ideally, $Y_{\rm else}$ should contain all accessible information complementary to already learned representations. However, in practice, the $Y_{\rm else}$ we get via auto-encoding is a lossy compression based on the latent feature salience with respect to the auto-encoding lost function. Information needed in a following task may already be compressed away since which information will be useful in the future is unknown. Allowing $Y_{\rm else}$ to be re-trained gives a chance for the lost information to be recovered.

\paragraph{Changes in input statistics.} The current framework requires that input space $X$ stays the same for all the following tasks to maximize interpretation. If $X$ are completely different, those tasks must be solved separately. What would happen if $X$ is slightly different? How to handle the situation depends on the strategy. For example, if the model working on the CLEVR dataset encounters a new shape: ``cone," following the current framework, the model would first classify it as a ``cylinder" until the model comes across some task that needs to tell apart ``cone" from ``cylinder." Then the model would pick some extra information from an un-conceptualized part like ``sharp top" to help distinguish ``cone" from ``cylinder." As a result, the model would think ``cone" is a sub-class of ``cylinder" with ``sharp top" and can further imagine a new shape like ``square" with ``sharp top," which is a ``square cone." Another example is when the distribution partially changes. Let's imagine, with the CLEVR dataset, a change where all balls suddenly become red. Under this situation,  the color and shape representation still works as before. However, since once independent representation color and shape now become dependent, interpretation for the following task now has ambiguity due to the redundancy.

\paragraph{Relationship with the biological brain.} The interpretation as meta-information is related to meta-cognition in the brain \citep{tobias2002knowing}. Especially, the un-conceptualized information map $Y_{else}$ is related to the meta-cognition aspect ``knowing what you do not know," which is very important for the proposed interpretation framework. Brain development study also supports the idea of progressive learning, with the most famous example being the critical period hypothesis \citep{lenneberg1967biological,toyoizumi2013theory}. Our interpretation framework is explicitly introducing independence among information maps. Meanwhile, there exist clues about the brain capable of performing independent component analysis only using local information available in each neuron \citep{isomura2016local}. Whether the brain is actively making use of this capability for task interpretation is not yet clear.  

\subsection{Limitations}
\label{sec:8-10}

We define the scope of the paper as a proposal of a new theoretical framework for progressive interpretation with simple proof-of-concept examples. This framework is still in its preliminary stage with limitations when considering practical implementation, which we will discuss as follows.

\textbf{Independence induction with IB:} The framework assumes that the information about the world can be divided into mutually independent partitions. Does this assumption really holds in the real world can be in question. However, there is no guarantee that our proposed IB-based method can find the right information map partition practically. As shown in Fig \ref{autoscan_info}, instead of an ideal black-or-white information separation, in practice, our IB based-method has to balance separation quality and task performance. This limitation is related to the point ``relationship between interpretation quality and performance." in the previous discussion section.

\textbf{About tasks order.} It should be pointed out that, the interpretation generated via our framework is sensitive to the order of the tasks. Then the natural question to ask is, what is the optimum task sequence. The authors believe that it is still an open question and an exciting research direction since we know the order of the task sequence matters not only for humans but also for machines \citep{bengio2009curriculum}. On the other hand, it is also possible that the model's interpretability can be boosted by carefully ordering tasks during learning. For example, using the example of CLEVR, one could explore the task order, where the model learns about ``red cube/sphere/cylinder" first and then tries to tell ``red". In this case, the model should use all information from the previous three tasks and learn to perform the ``or" operation. As a future direction,  guidelines on how to decide the optimal order can be a useful addition to the current work.

\textbf{Lack of suitable dataset:} The framework requires a sequence of mutually related tasks to generate useful interpretation. The lack of datasets organized in a sequentially organized manner is also a big issue limiting the practical usefulness of this framework. This problem can be mitigated either by developing new datasets with a clear curriculum design, such as textbook-like datasets, or by utilizing other techniques such as unsupervised sub-task generation \citep{kulkarni2016hierarchical}.

\textbf{Unchangeable inner representations:} In our framework, we assume that the inner representations, once formed, will not change over time. This assumption is clearly a simplification since, in reality, people's inner representations of the world are always subject to change. As a future direction, a mechanism is needed to constantly update the formed representations, delete inefficient ones, propose new ones, or even create a new level of representations by combining existing ones.

\bibliographystyle{APA}
\bibliography{ProgressiveInterpretation}

\end{document}